\documentclass[lettersize,journal]{IEEEtran}
\usepackage{amsmath,amsfonts}
\usepackage[ruled,linesnumbered]{algorithm2e}
\usepackage{changes}
\usepackage{array}
\usepackage{textcomp}
\usepackage{stfloats}
\usepackage{url}
\usepackage{xcolor}
\usepackage{verbatim}
\usepackage{graphicx}
\usepackage{cite}
\usepackage{amssymb}
\usepackage{listings}
\usepackage{enumerate}
\usepackage{multirow}
\usepackage[switch,mathlines]{lineno}
\usepackage{caption}
\usepackage[pagebackref=False,breaklinks=true,letterpaper=true,colorlinks,linkcolor=blue,anchorcolor=blue,citecolor=blue,urlcolor=blue,bookmarks=false]{hyperref}
\usepackage{doi}
\usepackage{lipsum}
\usepackage{pdfpages}

\usepackage{float}
\usepackage[caption=false,font=footnotesize,labelfont=rm,textfont=rm]{subfig}
\usepackage{booktabs}

\newcommand{\ie}{{i}.{e}., }

\newcommand{\etal}{\textit{~et~al}. }

\newcommand\figref[1]{Figure~\ref{#1}}
\newcommand\tableref[1]{Table~\ref{#1}}

\begin{document}

\title{A Multi-objective Evolutionary Algorithm Based on Bi-population with Uniform Sampling for Neural Architecture Search}

\author{
    Yu Xue,~\IEEEmembership{Senior Member,~IEEE}, 
    Pengcheng Jiang,~\IEEEmembership{Graduate~Student~Member,~IEEE}, \\Chenchen Zhu, Yong~Zhang,~\IEEEmembership{Senior~Member,~IEEE}, 
    Ran~Cheng,~\IEEEmembership{Senior~Member,~IEEE}, 
    Kaizhou~Gao,~\IEEEmembership{Senior~Member,~IEEE}, 
    Dunwei~Gong,~\IEEEmembership{Senior~Member,~IEEE}  
	\thanks{This work was supported by the National Natural Science Foundation of China (NO. 62376127, NO. 61876089, NO. 61876185), the Guangdong Basic and Applied Basic Research Foundation (No. 2024B1515020019), and the Natural Science Foundation of Shandong Province (NO. ZR2023ZD06). \textit{(Corresponding author: Yu Xue.)}}
	\thanks{Yu Xue, Pengcheng Jiang and Chenchen Zhu are with the School of Software, Nanjing University of Information Science and Technology, Nanjing 210044, China (e-mails: xueyu@nuist.edu.cn; pcjiang@nuist.edu.cn; 202212490283@nuist.edu.cn).}
    \thanks{Yong Zhang is with the School of Information and Control Engineering, China University of Mining and Technology, Xuzhou 221008, China (e-mail: yongzh401@cumt.edu.cn).}
    \thanks{Ran Cheng is with the Department of Data Science and Artificial Intelligence, and the Department of Computing, The Hong Kong Polytechnic University, Hong Kong SAR, China, and also with The Hong Kong Polytechnic University Shenzhen Research Institute, Shenzhen 518057, China (e-mail: ranchengcn@gmail.com).}
    \thanks{Kaizhou Gao is with the Macau Institute of Systems Engineering, Macau University of Science and Technology, Taipa 999078, Macao SAR, China (e-mail: kzgao@must.edu.mo).}
    \thanks{Dunwei Gong is with the College of Automation and Electronic Engineering, Qingdao University of Science and Technology, Qingdao 266061, Shandong, China (e-mail: dwgong@qust.edu.cn).}
}

\markboth{Journal of \LaTeX\ Class Files,~Vol.~14, No.~8, August~2021}%
{Shell \MakeLowercase{\textit{et al.}}: A Sample Article Using IEEEtran.cls for IEEE Journals}

\maketitle


\begin{abstract}
Neural architecture search (NAS) automates neural network design, improving efficiency over manual approaches. However, efficiently discovering high-performance neural network architectures that simultaneously optimize multiple objectives remains a significant challenge in NAS. Existing methods often suffer from limited population diversity and inadequate exploration of the search space, particularly in regions with extreme complexity values. To address these challenges, we propose MOEA-BUS, an innovative multi-objective evolutionary algorithm based on bi-population with uniform sampling for neural architecture search, aimed at simultaneously optimizing both accuracy and network complexity. In MOEA-BUS, a novel uniform sampling method is proposed to initialize the population, ensuring that architectures are distributed uniformly across the objective space. Furthermore, to enhance exploration, we deploy a bi-population framework where two populations evolve synergistically, facilitating comprehensive search space coverage.
Experiments on CIFAR-10 and ImageNet demonstrate MOEA-BUS's superiority, achieving top-1 accuracies of 98.39\% on CIFAR-10, and 80.03\% on ImageNet.
Notably, it achieves 78.28\% accuracy on ImageNet with only 446M MAdds.
Ablation studies confirm that both uniform sampling and bi-population mechanisms enhance population diversity and performance. Additionally, in terms of the Kendall's tau coefficient, the SVM achieves an improvement of at least 0.035 compared to the other three commonly used machine learning models, and uniform sampling provided an enhancement of approximately 0.07.
\end{abstract}


\section{Introduction}\label{sec:intro}
\IEEEPARstart{D}{eep} neural networks (DNNs) have achieved remarkable success in various fields, such as image and speech recognition~\cite{picco_deep_2024}, natural language processing~\cite{yang_instructtts_2024}, autonomous driving~\cite{gao_enhance_2024}, game and robotics~\cite{guo_pursuit-evasion_2024}, etc. With further research, DNNs are continuously optimized and improved, and their performance mainly depends on the structures of networks~\cite{sun_automatically_2020}. Traditional neural network architectures are usually designed manually by experts with extensive domain knowledge. Over time, these manually designed approaches have gradually shown limitations, especially when dealing with complex and high-dimensional data. Furthermore, as the size of datasets grows and computational resources increase, the demand for designing deeper and more complex networks increases~\cite{lee_az-nas_2024}. In this context, neural architecture search (NAS) has emerged, which aims to use algorithms to search for optimal network architectures, thus reducing human intervention and improving design efficiency~\cite{zoph_neural_2017}. Neural architecture search can not only optimize existing network architectures, but also explore new network architectures through the search process. These new architectures offer enhanced performance and higher generalization ability, thereby promoting the development and application of deep learning in various fields~\cite{wang_multi-population_2026}. The research and development of neural architecture search is of great significance in areas such as real-world applications and industrial production~\cite{rao_fx-darts_2025,liang_evolutionary_2025,tao_automatic_2024,yan_neural_2024,gambella_nachos_2025}.

Despite the significant progress made by neural architecture search in automating the design of neural network architectures, it still faces several challenges, including the scale of the search space, search efficiency, and model size constraints~\cite{yang_evolutionary_2024}. Existing NAS methods usually concern themselves only with the maximization of the classification accuracy~\cite{ding_stacked_2023, ding_bnas-v2_2022}. However, real-world applications often require neural networks to achieve a balance across multiple aspects. For example, models deployed on mobile devices need to maintain high accuracy while having a smaller model size and fast inference speed~\cite{lyu_efficient_2024}. With the widespread application of artificial intelligence technologies, the demand for efficient and high-performance models is increasing, which has prompted researchers to explore neural network architectures that can meet multiple performance needs. Therefore, some researchers have begun to conduct in-depth research on multi-objective neural architecture search, attempting to find architectures that can take into account multiple performance indicators~\cite{garcia-garcia_cgp-nas_2022}. Unlike single-objective optimization, multi-objective optimization requires considering multiple performance indicators at the same time, which usually means finding the balance among these indicators, rather than a single optimal solution~\cite{li_multiobjective_2025}. Evolutionary algorithms, by simulating natural selection and genetic mechanisms, maintain a population of candidate solutions and improve these solutions through operations such as selection, crossover, and mutation in each generation~\cite{dong_cell-based_2023}. Evolutionary algorithms have good global search capabilities and can flexibly and effectively explore and handle Pareto optimization in multi-objective space. In contrast, the two other popular categories of NAS methods: reinforcement learning-based (RL)~\cite{pham_efficient_2018} and gradient-based (GD)~\cite{ding_nap_2022} methods, have some limitations when dealing with multi-objective problems. Gradient-based methods, such as DARTS~\cite{liu_darts_2018}, usually assume that the optimization problem is differentiable and has only one objective function. However, some indicators of architectures, such as model complexity, are usually non-differentiable and cannot be easily optimized through the loss function. In addition, gradient-based methods may tend to optimize the objectives that contribute the most to the gradient signal, while neglecting other equally important objectives~\cite{cai_sto-darts_2024}. Reinforcement learning-based methods usually rely on a reward function to guide the search process, but in the case of multi-objective, defining a reward function that fully reflects all objectives is very difficult~\cite{lin_bandit-nas_2024}. Moreover, they consume more computational resources and incur higher time costs than the other two methods~\cite{tan_mnasnet_2019}. Overall, evolutionary algorithms are more suitable for multi-objective neural structure search, as they provide an effective search strategy.
In current NAS methods, some research employs multi-objective optimization theory to simultaneously optimize multiple metrics, with network complexity being a common second metric besides classification accuracy. The frequently used approaches to represent network complexity include the number of parameters in the network or ``multiplying and accumulating operations (MAdds)''.

In multi-objective evolutionary optimization methods, population diversity determines the distribution of the population on the Pareto front. A population lacking diversity tends to converge to one or more regions in the objective space while neglecting other parts. During the evolutionary process, a population with insufficient diversity tends to focus solely on exploiting known regions of the objective space, thereby neglecting the exploration of new areas. This results in a final solution set where the trade-off solutions are not representative across each objective. In multi-objective evolutionary neural architecture search (MO-ENAS), this issue is often overlooked. For instance, NSGA-Net focuses more on architectures around a specific MAdds value, resulting in a population that lacks diversity, limiting the breadth of search, and leading to architectures that are locally optimal in this region. Based on analysis of this problem, population initialization and selection operators during the search process are identified as two critical factors. 
In the objective space of NAS, medium-sized architectures often have a large number of different representations of encoding, but small and large architectures do not.
Therefore, commonly used random initialization is not entirely suitable for the NAS search spaces, which leads to a bias toward small and medium-sized network architectures in terms of MAdds during population initialization. Additionally, relying solely on non-dominated sorting-based selection operators makes it difficult to maintain good population diversity during the search process. Multi-population mechanisms are common, flexible, and effective methods for enhancing population diversity. Under existing selection operators, multi-population mechanisms can significantly improve population diversity.

Another key challenge in NAS stems from the substantial resources consumed in evaluating numerous candidate architectures. Although there are currently many studies on training-free evaluation, they still do not have significant advantages compared to traditional evaluation acceleration methods~\cite{yamasaki_rbflex-nas_2025}. Therefore, during the search process, each architecture requires training to obtain accuracy for environmental selection, which consumes considerable resources and requires extensive time. To address this issue, ENAS methods commonly employ surrogate models, weight inheritance, and other techniques. Weight inheritance methods aim to utilize pre-trained weights obtained from supernets to initialize parameters of identical modules in architectures, thereby reducing training time for individual architectures. This approach can significantly shorten the search duration of the original algorithm. However, architectures still require at least one inference time for actual evaluation even when using one-shot methods, which prevents a large number of architectures from being searched. Surrogate models reduce the number of architectures requiring actual evaluation by predicting architecture performance. The resource and time consumption of this prediction process are substantially lower than the inference cost of network architectures, thus enabling rapid evaluation of numerous architectures during the search process.

To address the above problems, we propose an effective algorithm, called MOEA-BUS, a multi-objective evolutionary algorithm based on bi-population with uniform sampling for neural architecture search. Firstly, we design a uniform sampling method for initializing the population so that the initial architectures are distributed as uniformly as possible in the objective space. Second, to explore the search space more fully during the search process, we propose a multi-objective bi-population-based evolutionary algorithm where two populations evolve concurrently and exchange individuals. The proposed method aims to provide a set of high-performance architectures that take into account multiple optimization objectives. We validate the effectiveness of the proposed algorithm on an image classification task using the standard datasets CIFAR-10, CIFAR-100, and ImageNet. The computational results show that the proposed method outperforms most state-of-the-art NAS methods. In addition, we conduct sufficient ablation studies for each key mechanism to prove the effectiveness of the proposed method. The main contributions are as follows:
\begin{enumerate}[1)]
    \item The proposed method simultaneously optimizes accuracy and network complexity, with MAdds as the complexity metric. During the search process, a surrogate model and weight inheritance are used to reduce the time and resources required to evaluate the architectures.
    \item Uniform sampling is proposed to improve the quality of the initial population, in which a two-stage sampling method is designed to sample individuals and initialize an initial population that is uniform on the network complexity, \ie MAdds.
    \item A multi-objective bi-population-based evolutionary algorithm is proposed, in which two populations evolve together and genes are exchanged between them to fully explore the search space. It can largely prevent the algorithm from falling into a local optimum while accelerating convergence.
\end{enumerate}

The remainder of this paper is organized as follows: Section \ref{sec:RelatedWork} presents related work and background. Section \ref{sec:method} describes the proposed method in detail. We present the experimental design to verify the effectiveness and efficiency of the proposed method and discuss the results in Section \ref{sec:experiment}. Finally, conclusions and future work are outlined in Section \ref{sec:conclusion}.

\section{Related Work}\label{sec:RelatedWork}

\subsection{Multi-objective NAS}\label{sec:RelatedWork:multi-objectiveNAS}
Existing research in NAS concentrates mainly on improving the accuracy of neural networks, but these single-objective methods often ignore the more complex requirements of real-world applications. Although these existing networks perform well on recognition tasks, they are often difficult to deploy in real-world situations due to high computational costs and large model sizes. 
Researchers have turned to multi-objective optimization for NAS and explore how to more effectively find the optimal balance between these metrics to design neural network architectures that are both efficient and practical. 
For example, Lu\etal use NSGA-II as the multi-objective optimization method to simultaneously optimize accuracy and computational cost~\cite{lu_multiobjective_2021,lu_nsga-net_2019}. Subsequently, they further investigate methods to reduce the time consumption of multi-objective optimization by introducing a surrogate model~\cite{lu_nsganetv2_2020,lu_surrogate-assisted_2022}. In addition, Xue\etal propose a multi-objective evolutionary algorithm for NAS that focuses on accuracy and time consumption~\cite{xue_neural_2023}. Wang\etal improve the particle swarm optimization (PSO) algorithm to optimize both classification accuracy and MAdds~\cite{wang_evolving_2019}. Du\etal design an environmental selection operation based on reference points to improve the multi-objective optimization process in NAS~\cite{tong_neural_2022}. 
Although these studies have yielded successful results in multi-objective optimization, they usually require evaluation of a large number of architectures, which is time-consuming and inefficient. In addition, among the existing multi-objective NAS methods, there are relatively few studies and improvements on multi-objective evolutionary algorithms, and researchers tend to choose only off-the-shelf algorithms, such as NSGA-II, to handle multi-objective optimization problems in NAS. Therefore, an improved multi-objective algorithm is proposed in order to better adapt to the search framework in this work.

\subsection{Multi-population ENAS}\label{sec:RelatedWork:multi-pop-NAS}
Multi-population strategies in ENAS are designed to enhance search diversity and prevent premature convergence. However, these methods encounter a fundamental paradox: while the migration of high-performing individuals between populations is intended to share beneficial traits, it can inadvertently homogenize the gene pool, ultimately converging to a single suboptimal solution. To address this issue, recent research has proposed more sophisticated strategies. These include creating heterogeneity by employing different evolutionary algorithms~\cite{wang_multi-population_2026}, implementing intelligent migration protocols that select for novelty to increase diversity~\cite{xue_pairwise_2026}, and redefining the search to evolve functionally specialized networks that are combined for superior performance~\cite{song_multi-population_2024}.
However, these methods do not further explore the lack of population diversity caused by the uneven distribution of objective space in NAS, nor do they make adjustments according to this characteristic.

\subsection{Diversity Preservation}\label{sec:RelatedWork:surrogatemodels}
In evolutionary multi-objective algorithms, preserving population diversity is crucial for helping the algorithm to avoid falling into a local optimum and explore a globally optimal solution, and many scholars have conducted research to balance diversity and convergence.
Saad\etal propose a multi-objective artificial bee colony (ABC) algorithm~\cite{saad_multi-objective_2018}. The algorithm relies on the basic principle of population evolution, which exploits the differences among individuals in the population to generate new candidate solutions, effectively making use of the diversity among individuals and promoting the evolution of the whole population. Wang\etal combine the differential evolution algorithm with the particle swarm optimization, which uses an adaptive mutation strategy, achieving effective preservation of population diversity at the early stage and significantly accelerating the convergence rate at the later stage during the evolution~\cite{wang_self-adaptive_2019}. It can be seen that designing better search strategies can accelerate convergence speed, improve population diversity, and enhance effective interactions between individuals, thereby ultimately enhancing the performance of the multi-objective evolutionary algorithms (MOEAs). 
Therefore, careful consideration and design of appropriate search strategies are crucial for obtaining satisfactory results~\cite{tang_differential_2015}. In addition, initialization methods can be adjusted to integrate external information at the outset in the population initialization phase of multi-objective evolutionary algorithms, aiming to approximate the global optimum solution as closely as possible~\cite{kazimipour_review_2014}. Evolutionary strategies are crucial for MOEAs to rapidly converge to the Pareto front. Thus, we design a multi-objective evolutionary algorithm for NAS from two perspectives of the initialization and search strategy.

\section{Proposed Method For Multi-objective Evolutionary Neural Architecture Search}\label{sec:method}
This section presents the details of a bi-population-based multi-objective evolutionary algorithm with uniform sampling for NAS. We firstly present the framework of the proposed algorithm in Section \ref{sec:method:framework}. Then, the details of the proposed search space and encoding are introduced in Section \ref{sec:method:encoding}. Subsequently, the proposed uniform sampling method is described in Section \ref{sec:method:sampling}, and proposed multi-objective algorithm with bi-population is described in Section \ref{sec:method:MOEA-BUS}. Finally, surrogate model and the use of supernet are introduced in Section \ref{sec:method:efficiency}.

\begin{figure*}[t]
    \centering
    \includegraphics[width=0.9\textwidth]{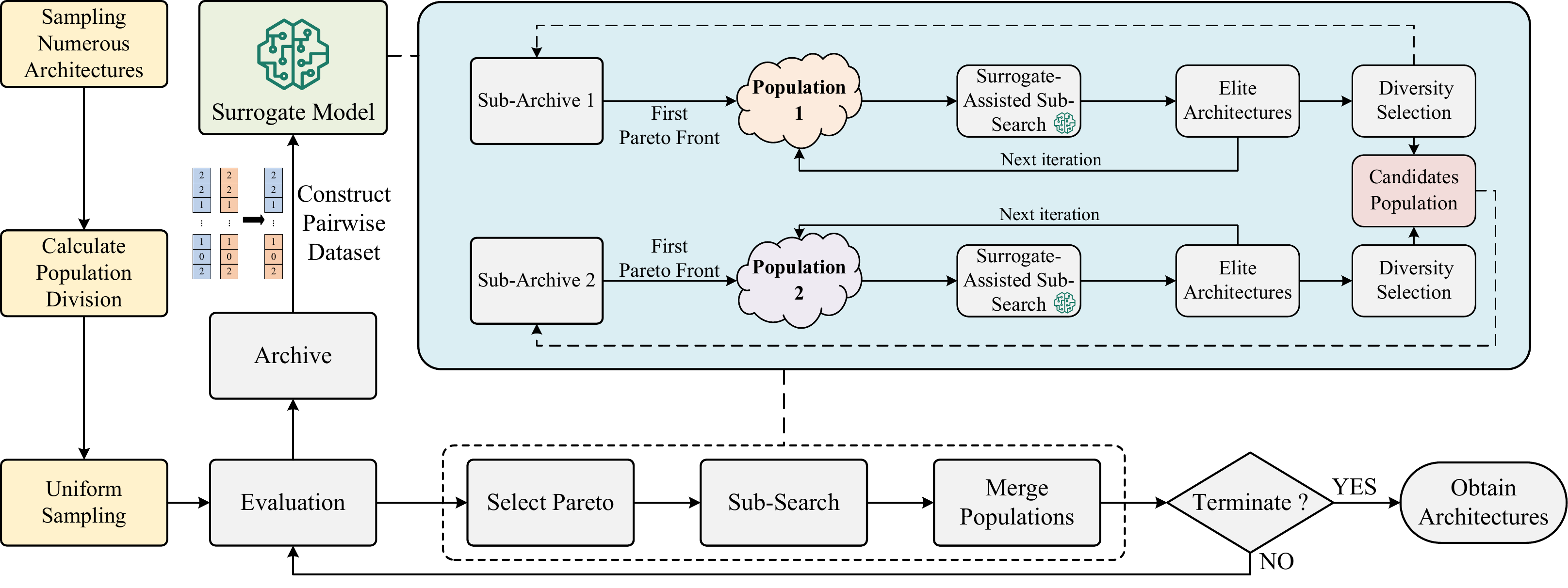}
    \caption{\textbf{Overall Framework:} A multi-objective evolutionary neural architecture search method based on bi-population with uniform sampling.
    }
    \label{fig:flowchat}
\end{figure*}

\subsection{Overall Framework}\label{sec:method:framework}
The existing multi-objective evolutionary neural architecture search methods are prone to the problem of lack of diversity due to conflicting objectives, and the proposed method suggests two improvement measures: firstly, a uniform sampling method is designed to initialize the initial population; secondly, two populations jointly perform evolutionary exploration of the search space to improve population diversity during the search. An overview of the proposed overall framework is illustrated in Fig. \ref{fig:flowchat}.



First, a large number of architectures are sampled and their MAdds is evaluated, with uniform sampling being used to obtain candidate architectures that are uniformly distributed across the MAdds metric. These selected architectures serve as the initial architectures of archive $\mathcal{A}$ and undergo real evaluation. Subsequently, these architectures are divided into two archives, one containing medium-sized architectures and another containing large and small architectures. These two archives are respectively used for the search processes of two populations.
The core idea of the uniform sampling method is to ensure that the individuals in the initial population are uniformly distributed in the objective space, avoiding architecture concentration in certain regions and improving search space coverage. Uniform sampling helps enhance the diversity and global exploration ability in the early search phase. To strengthen information exchange between populations and solution diversity, the proposed method performs an exchange of individuals between populations at the end of each generation.  Population 1 shares excellent elite individuals with population 2, thereby promoting comprehensive search space coverage and diversity maintenance, and accelerating the convergence of the entire search process. Meanwhile, the computational cost from the evaluation during the search is reduced with the help of a surrogate model and weight inheritance technique. After several generations, all the searched network architectures are sorted by non-dominated sorting and a set of high quality architectures are chosen based on specific preferences.

\begin{figure*}[t]
    \centering
    \includegraphics[width=0.75\textwidth]{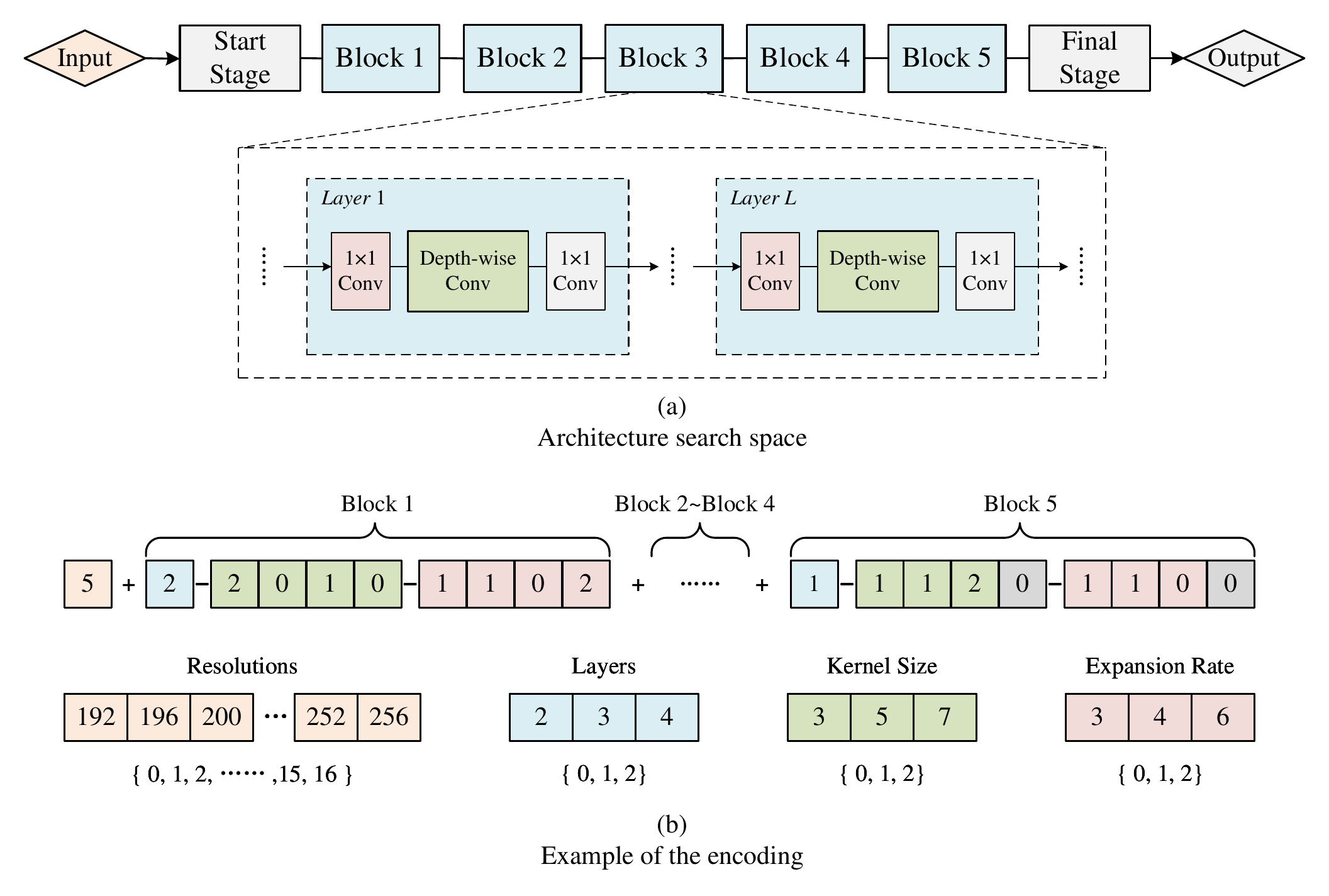}
    \caption{Search space and encoding. (a) The architecture search space. (b) An example of the encoding. The encoding is divided into five parts by blocks. The parameters we search include image resolution, the number of layers in each block, the expansion rate, and the kernel size in each layer.}
    \label{fig:encoding}
\end{figure*} 
\subsection{Search Space and Encoding}\label{sec:method:encoding}
The quality of evolutionary search results is fundamentally determined by the chosen search space. In this work, architectures are based on MobileNetV3~\cite{howard_searching_2019} and are composed of three stages. The initial stage and final stage remain fixed. The main part of architectures consists of a stack of multiple convolutional blocks. Externally, the size of the input image (resolution) also needs to be searched. In the internal structure, each block contains several layers, and the numbers of layers are optional. In addition, each layer uses an inverted bottleneck structure that contains multiple convolutions, requiring optimization of both convolution kernel size and expansion rate. Fig. \ref{fig:encoding} illustrates the search space and encoding strategy. The algorithm searches for the appropriate expansion ratios for the initial $1\times1$ convolution and kernel sizes for the depth-wise separable convolution in each layer. The encoding of an architecture is composed of image resolution and other parts representing five blocks. Each block's encoding specifies the number of layers, expansion rate, and kernel size of its constituent convolution layers. The encoding's values correspond to indices from predefined considered option sets. Moreover, the absence of a layer is indicated by a padded zero to achieve the fixed length encoding, which is not from considered options.

\begin{figure}[h]
    \centering
    \includegraphics[width=0.8\linewidth]{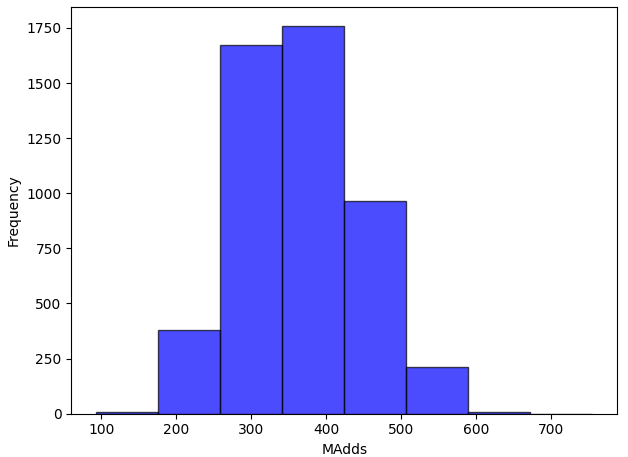}
    \caption{The distribution of randomly sampled 5,000 architectures.}
    \label{fig:random5000}
\end{figure}

\begin{figure}[h]
    \centering
    \includegraphics[width=\linewidth]{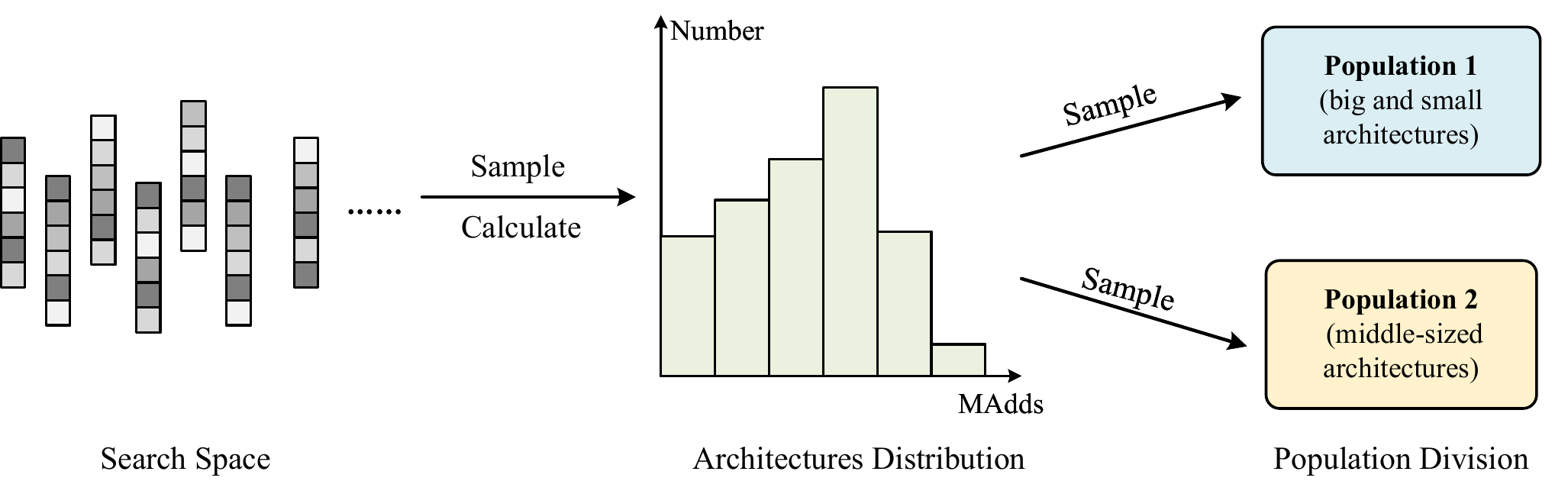}
    \caption{The illustration of uniform sampling.}
    \label{fig:uniform-sample}
\end{figure}

\subsection{Uniform Sampling}\label{sec:method:sampling}
During the evolutionary process, the selection and distribution of the initial population critically determines both the search efficacy of the method and the performance of the surrogate model. A well-designed initial population provides diverse starting points that enhance the global search capability, while a poor initial population may lead to the search falling into local optimum and limit the exploration of the search space.
Furthermore, the initial archive derived from a uniformly distributed initial population proves beneficial for surrogate model training, enabling more precise identification of superior architectures in subsequent search iterations.

To investigate this phenomenon, we sampled 5,000 architectures from the search space using random sampling method and analyzed their distributional characteristics. Fig.~\ref{fig:random5000} illustrates the distribution of these 5000 architectures across the metric of network complexity. The horizontal coordinate is the MAdds metric, and the vertical coordinate is the count of architectures. As can be seen in Fig.~\ref{fig:random5000}, randomly sampled architectures exhibit a highly concentrated distribution pattern on the MAdds metric, with the overwhelming majority clustering between 300M and 400M MAdds. This means that architectures within this complexity range occupy the majority of the search space, while architectures with higher or lower complexity remain relatively scarce. This concentrated distribution limits the capacity of the population to explore in regions of higher or lower complexity, resulting in that potentially valuable architectures may be overlooked at an early stage. Consequently, in subsequent evolutionary iterations, the evolutionary algorithm tends to generate new architectures that closely resemble the current population, further limiting architectural diversity and search effectiveness.

In order to obtain high-quality initial populations, we propose a uniform sampling method illustrated in Fig. \ref{fig:uniform-sample}. Specifically, the uniform sampling method proceeds as follows: Initially, a substantial number of architectures are randomly sampled from the search space, and their complexity (MAdds) is calculated. Subsequently, all architectures are sorted by MAdds values and divided into several regions with uniform ranges based on MAdds distribution according to their complexity from smallest to largest. After division, a certain number of architectures are selected from each region. In order to ensure diversity across the search space, regions with high and low MAdds values are emphasized, containing architectures with extreme complexity. Architectures from these extreme regions are selected and merged to form population 1. Meanwhile, architectures with moderate MAdds values are selected and merged to constitute population 2. These two initial populations ensure the diversity and provide a rich architectural pool for subsequent evolution. Through uniform sampling, the initial population covers multiple complexity regions, from low to high, achieving a more uniform distribution in the objective space.

\subsection{Multi-objective Evolutionary Algorithm Based on Bi-population}\label{sec:method:MOEA-BUS}
\begin{algorithm}[t]
\SetAlgoLined
\caption{Framework of MOEA-BUS}\label{algo:MOEA-BUS}
\KwIn{Supernet $W_s$, number of iterations $T$.}
$\mathcal{H} \leftarrow $ Initialize numerous architectures; \label{algo:MOEA-BUS:line:1}

$\mathcal{A} \leftarrow \varnothing $; \tcp{Create an empty archive for storing records.}

$P_1, P_2 \leftarrow $ Uniform\_Sampling($\mathcal{H}$); \tcp{The initial populations are constructed using the proposed uniform sampling method. See Section \ref{sec:method:sampling} for details.} \label{algo:MOEA-BUS:line:3}

\For{$a$ in $P_1\cup P_2$}{\label{algo:MOEA-BUS:line:4}

$W_a \leftarrow W_s(a)$; \tcp{Inherit the weights of corresponding pre-trained modules in the supernet according to architecture $a$.}

$error\_rate \leftarrow \text{SGD}(a,W_a)$;

$\mathcal{A} \leftarrow \mathcal{A} \cup \{(a, error\_rate)\}$;

}\label{algo:MOEA-BUS:line:8}

$t \leftarrow 0$;

\While{$t < T$}{

$predictor \leftarrow$ Construct surrogate model with $ \mathcal{A}$; \tcp{See Section \ref{sec:method:efficiency} for details.} \label{algo:MOEA-BUS:line:11}

$P^*_1 \leftarrow \text{Sub-Search}(P_1, predictor, \mathcal{A})$; \tcp{Search with the small and big architectures.} \label{algo:MOEA-BUS:line:12}

$P^*_2 \leftarrow \text{Sub-Search}(P_2, predictor, \mathcal{A})$; \tcp{Search with the middle-sized architectures.} \label{algo:MOEA-BUS:line:13}

\For{$a$ in $P^*_1\cup P^*_2$}{

Same as lines \ref{algo:MOEA-BUS:line:4} to \ref{algo:MOEA-BUS:line:8}; \tcp{Evaluate $a$ with supernet and add the records into $\mathcal{A}$}

}

\tcp{Update the initial populations for two Sub-Search processes.}

$P_1 \leftarrow P_1 \cup P^*_1$; \label{algo:MOEA-BUS:line:17}

$P_2 \leftarrow P_2 \cup P^*_1 \cup P^*_2$;  \label{algo:MOEA-BUS:line:18}

$t \leftarrow t+1$;

}

\Return Final population (all individuals in $\mathcal{A}$).
\end{algorithm}

\begin{algorithm}[t]
\SetAlgoLined
\caption{Sub-Search ($Pop$, $predictor$, $\mathcal{A}$)}\label{algo:Sub-Search}
\KwIn{Number of generations $G$.}
$g \leftarrow 0$;

$P \leftarrow $ Get first Pareto front of $Pop$; \tcp{Only use the first Pareto front as the initial population.} \label{algo:Sub-Search:line:2}

$P.F_1 \leftarrow $ Predict the strength for each architecture with $predictor$; \tcp{See Section \ref{sec:method:efficiency} for details.} \label{algo:Sub-Search:line:3}

$P.F_2 \leftarrow $ Calculate the MAdds of each architecture; \label{algo:Sub-Search:line:4}

\While{$g<G$}{

$Q \leftarrow$ Generate offspring of $P$ with crossover and mutation;

$P.F_1, Q.F_1 \leftarrow $ Predict the strength for each architecture in $P \cup Q$ with $predictor$; \tcp{See Section \ref{sec:method:efficiency} for details.}

$Q.F_2 \leftarrow $ Calculate the MAdds of each architecture;

$P \leftarrow P \cup Q$;

$P \leftarrow$ Non-Dominated-Sort($P$);

$P \leftarrow$ Crowded-Selection($P$);

$g \leftarrow g+1$;
}

$PF \leftarrow$ Non-Dominated-Sort($P$);

\Return Diversity\_Selection($FP$, $\mathcal{A}$).
\end{algorithm}

In order to further increase the population diversity and explore the search space more comprehensively, a bi-population evolution framework is proposed. The core idea of the proposed framework is to introduce two populations for parallel evolution to increase the diversity of solutions during the search process and to improve the global search capability of the algorithm. The bi-population evolution process is the main loop part of the evolutionary search.

Algorithm \ref{algo:MOEA-BUS} demonstrates the overall process of bi-population search. First, two populations are obtained according to the uniform sampling method in Section \ref{sec:method:sampling}, where population 1 contains large and small architectures, and population 2 contains medium-sized architectures (lines \ref{algo:MOEA-BUS:line:1}-\ref{algo:MOEA-BUS:line:3}). Subsequently, all individuals in both populations are truly evaluated, and the results of real evaluation are recorded in archive $\mathcal{A}$ (lines \ref{algo:MOEA-BUS:line:4}-\ref{algo:MOEA-BUS:line:8}). During this process, each architecture needs to be trained and validated using image datasets, and the classification error rate on the validation set is obtained. Since the training and validation of architectures in the overall population are mutually independent processes, the real evaluation of each architecture is divided into sub-tasks that are automatically allocated to multiple available GPUs by a single device for execution. Afterwards, we set $T$ rounds of iterative search while continuously updating the overall archive and training the surrogate model. Specifically, we first train a surrogate model based on the current overall archive $\mathcal{A}$ (line \ref{algo:MOEA-BUS:line:11}), then use the surrogate model to assist the respective evolutionary processes of the two populations (lines \ref{algo:MOEA-BUS:line:12}-\ref{algo:MOEA-BUS:line:13}). The evolutionary process of each population is detailed in Algorithm \ref{algo:Sub-Search}. Subsequently, the two elite populations ($P_1^*$ and $P_2^*$) obtained from the search undergo real evaluation, following the same process as lines \ref{algo:MOEA-BUS:line:4} to \ref{algo:MOEA-BUS:line:8}. Afterwards, we add individuals from elite population $P_1^*$ to initial populations $P_1$ and $P_2$, and add individuals from elite population $P_2^*$ only to initial population $P_2$ (lines \ref{algo:MOEA-BUS:line:17}-\ref{algo:MOEA-BUS:line:18}). When $T$ rounds of iterations are satisfied, the algorithm terminates, and the final overall population (i.e. all individuals in $\mathcal{A}$) are obtained.

Algorithm \ref{algo:Sub-Search} presents the detailed process of sub-search (lines \ref{algo:MOEA-BUS:line:12}-\ref{algo:MOEA-BUS:line:13} in Algorithm \ref{algo:MOEA-BUS}). First, based on the current sub-population, non-dominated sorting is performed and the Pareto front is obtained, where individuals in the first rank serve as the initial population $P$ for sub-search (line \ref{algo:Sub-Search:line:2}). During the surrogate evaluation process for $P$, the surrogate model is used to predict the strength of each individual as the fitness value for the first objective ($P.F_1$), while the computational complexity MAdds of each architecture individual in $P$ is calculated as the fitness value for the second objective ($P.F_2$). Subsequently, $G$ generations of search are executed, where new individuals are generated through crossover and mutation operations in each iteration, and the fitness values of offspring ($Q.F_1$ and $Q.F_2$) are obtained through the same surrogate evaluation process as in lines \ref{algo:Sub-Search:line:3}-\ref{algo:Sub-Search:line:4}. The only difference is that $Q.F_1$ is calculated using both $P$ and $Q$, so $P.F_1$ is also updated. Then, non-dominated sorting and crowding distance selection are performed on the combined population to obtain the next generation population $P$. After $G$ generations of search, diversity selection is applied to the final population $P$ that has undergone non-dominated sorting to obtain the architecture most different from the overall population $\mathcal{A}$. The specific operation of diversity selection involves computing the fitness value differences between individuals in $FP$ and each individual in the overall population layer by layer, and retaining the individual with the maximum difference.

\begin{figure}[t]
    \centering
    \includegraphics[width=\linewidth]{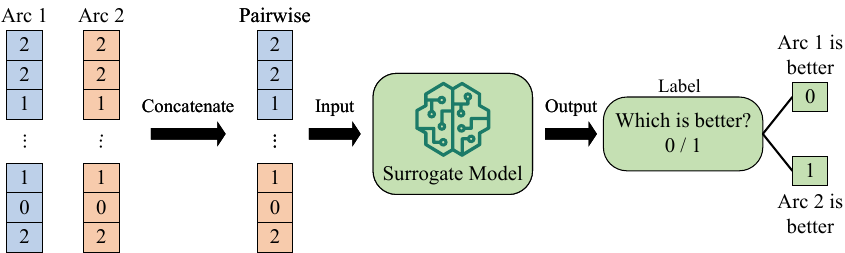}
    \caption{The illustration of the proposed surrogate model.}
    \label{fig:surrogate_process}
\end{figure}
\subsection{Surrogate-assisted Search and Weight Inheritance}\label{sec:method:efficiency}
A key challenge in the field of NAS lies in the substantial computational overhead required to evaluate numerous network architectures~\cite{zhang_gpu_2023}. This problem is prevalent despite the fact that different search spaces and strategies are used. To improve the search efficiency of the proposed method, we use a surrogate model for sub-search process and weight inheritance for real evaluation to reduce the search time.
We construct a surrogate model used to predict performance ranking of architectures during the sub-search process, which costs much less time for evaluation and is able to identify the potential architectures.
In this paper, support vector machine (SVM) is chosen to build the surrogate model based on comparison relationships between architectures.
First, we concatenate the encoding of each individual in the archive with the encoding of every individual numbered after it to construct training data D, and set the data labels to 0 or 1, where 0 indicates that the preceding architecture in the concatenated encoding is better, and 1 indicates that the following architecture is better. This process is illustrated in Fig.~\ref{fig:surrogate_process}. Subsequently, this dataset is used to train a fitted SVM model. During the prediction phase, we apply the same processing to the architectures that need to be predicted. Assuming the prediction result for $(X_i, X_j)$ is $Pred$, then the $i$-th architecture and the $j$-th architecture will each receive a score, where the $i$-th architecture obtains a score of $Pred$, and the $j$-th architecture obtains a score of $(1-Pred)$. Finally, we use the accumulated scores as intensity, where higher intensity indicates higher classification error rate of the architecture.
The non-dominated sorting for architectures is performed by the predicted strength and the calculated MAdds.

Additionally, the construction and training of the surrogate model require already evaluated architectures to be used as training samples, and we use weight inheritance to accelerate the evaluation of architectures. When evaluating the performance of an architecture, the weights of Once-For-All~\cite{cai_once_2020} are used as initialization for the gradient descent algorithm, thereby significantly reducing the time of training and evaluation for candidate architectures. Once-For-All is a well-trained supernet built upon the MobileNetV3 backbone network, encompassing more than $10^{19}$ candidate sub-networks. In this paper, candidate networks directly inherit their weights from the supernet. During training process of them, the weights of candidate networks are updated, and the corresponding weights in the supernet remain frozen.

\section{Experiments}\label{sec:experiment}
In this section, we conduct a series of experiments to validate the effectiveness of the proposed algorithm. Initially, Section \ref{sec:experiment:configurations} describes the specific configuration of the experiments. Subsequently, we present and analyze the experimental results on the most commonly used datasets for image classification from different perspectives in Section \ref{sec:experiment:results}. Additionally, we conducted ablation experiments on two key mechanisms in the paper and analyze the results in Section \ref{sec:experiment:abla-key} to demonstrate the effectiveness of the proposed method.
Then, we discuss the surrogate model in Section \ref{sec:experiment:surrogate}. Furthermore, the effectiveness of the uniform sampling strategy is shown in Section \ref{sec:experiment:resultsample}. Finally, the ablation study on the bi-population mechanism and analysis are presented in Section \ref{sec:experiment:Bi-population}.

\subsection{Experimental Configurations}\label{sec:experiment:configurations}

We conduct experiments using three widely recognized image classification datasets: CIFAR-10, CIFAR-100, and ImageNet. CIFAR-10 consists of 60,000 32x32 color images across 10 classes. CIFAR-100 is similar but contains 100 classes. ImageNet contains over 1.2 million images belonging to 1,000 different classes. ImageNet is known for its vast variety of images and challenging classification tasks, making it a benchmark for evaluating deep learning models. The performance of architectures discovered by the proposed algorithm is evaluated based on accuracy and MAdds. MAdds provides insights into the computational complexity of architectures. The summary of parameter settings for MOEA-BUS is presented in \tableref{tab:cfgs}. Our experiments are conducted on a single RTX 3090 (24GB) card using PyTorch 2.0 and CUDA 11.7 environment. The code is available on \url{https://github.com/pcjiang1998/MOEA-BUS}.

\begin{table}[h]
    \centering
    \caption{Detailed settings of MOEA-BUS.}
    \label{tab:cfgs}
\begin{tabular}{l|l}
\hline
\multicolumn{2}{c}{\textbf{In main framework}}                                       \\ \hline
Size of initial population 1                            & 25                  \\
Size of initial population 2                            & 75                  \\
Number of iterations                                    & 25                  \\\hline\hline
\multicolumn{2}{c}{\textbf{In sub-search}}                                             \\\hline
Number of generations                                   & 40                  \\
Size of population                                      & 60                  \\
\multirow{2}{*}{Number   of reserved elite individuals} & 4 for population 1  \\
                                                        & 6 for population 2  \\
Crossover   method                                      & Two-point crossover \\
Mutation method                                         & Polynomial mutation \\\hline\hline
\multicolumn{2}{c}{\textbf{In final train}}                                            \\\hline
Number of epochs                                        & 200                 \\
Optimizer                                               & SGD                 \\
Initial learning rate                                   & 0.01                \\
Momentum rate                                           & 0.9                 \\
Batch size                                              & 128                 \\ \hline
\end{tabular}
\end{table}
\begin{table}[t]
    \centering
    \caption{Comparison on the CIFAR-10 dataset. This table compares the classification accuracy, computational complexity (MAdds), and search cost with other state-of-the-art NAS methods on the CIFAR-10 dataset.}
    \label{tab:cifar10_result}
    \scalebox{0.75}{
    \begin{tabular}{c|c|c|c|c|c|c}
    \hline\hline
        \textbf{Architecture} & \textbf{\begin{tabular}[c]{@{}c@{}}Accuracy\\(\%)\end{tabular}} & \textbf{\begin{tabular}[c]{@{}c@{}}MAdds\\(M)\end{tabular}} & \textbf{\begin{tabular}[c]{@{}c@{}}Params\\(M)\end{tabular}} & \textbf{\begin{tabular}[c]{@{}c@{}}Search Cost\\(GPU Days)\end{tabular}} & \textbf{\begin{tabular}[c]{@{}c@{}}Search\\Method\end{tabular}}  & \textbf{Year}\\ 
        \hline
        MobileNetV2~\cite{sandler_mobilenetv2_2018} & 95.74 & 300 & 2.2 & - & manual & 2018 \\ 
        EfficientNet-B0~\cite{tan_efficientnet_2019} & 98.1 & 387 & 4.0 & - & manual & 2019 \\ 
        \hline
                
        NASNet-A~\cite{zoph_learning_2018} & 97.35 & 608 & - & 1800 & RL & 2018 \\ 
        BNAS~\cite{ding_bnas_2021} & 97.03 & - & 4.7 & 0.19 & RL & 2021    \\
        DBNAS-B~\cite{yang_deeply_2025} & 97.33 & - & 3.1 & - & RL & 2025    \\
        \hline
        
        PC-DARTS~\cite{xu_pc-darts_2019} & 97.43 & 558 & 3.6 & 0.1 & GD & 2019 \\ 
        P-DARTS~\cite{chen_progressive_2019} & 97.5 & 532 & 3.4 & 0.3 & GD & 2019 \\ 
        FairDARTS~\cite{chu_fair_2020} & 97.46 & 373 & 2.8 & 0.25 & GD & 2020 \\ 
        NoisyDARTS~\cite{chu_noisy_2021} & 97.63 & 534 & 3.3 & 0.4 & GD & 2021 \\ 
        EoiNAS~\cite{zhou_exploiting_2022} & 97.50 & - & 3.4 & 0.6 & GD & 2022    \\
        iDARTS~\cite{wang_idarts_2023} & 97.47 & - & 3.6 & - & GD & 2023    \\
        SWD-NAS~\cite{xue_self-adaptive_2024} & 97.49 & 519 & 3.17 & 0.13 & GD & 2024\\
        PA-DARTS~\cite{xue_improved_2024} & 97.59 & 578 & 3.75 & 0.36 & GD & 2024\\
        GENAS~\cite{xue_gradient-guided_2024} & 97.55 & - & 3.53 & 0.26 & GD & 2024    \\
        DBNAS-A~\cite{yang_deeply_2025} & 97.58 & - & 2.4 & - & GD & 2025    \\
        DBNAS-C~\cite{yang_deeply_2025} & 97.50 & - & 2.9 & - & GD & 2025    \\
        FX-DARTS~\cite{rao_fx-darts_2025} & 95.96$\pm$0.01 & 195 & 1.26 & 0.11 & GD & 2025    \\
        \hline
        
        AmoebaNet-B~\cite{real_regularized_2019} & 97.5 & 555 & - & 3150 & EA & 2019 \\ 
        NSGA-Net~\cite{lu_nsga-net_2019} & 97.25 & 535 & 3.3 & 4 & EA & 2019 \\
        CARS~\cite{yang_cars_2020} & 97.43 & 728 & 3.6 & 0.4 & EA & 2020 \\ 
        FairNAS-A~\cite{chu_fairnas_2021} & 98.2 & 391 & - & 12 & EA & 2021 \\ 
        FairNAS-B~\cite{chu_fairnas_2021} & 98.1 & 348 & - & 12 & EA & 2021 \\ 
        FairNAS-C~\cite{chu_fairnas_2021} & 98.0 & 324 & - & 12 & EA & 2021 \\ 
        MPAE-A~\cite{zou_multiple_2024} & 97.35 & - & 2.8 & 0.3 & EA & 2024    \\
        MPAE-B~\cite{zou_multiple_2024} & 97.39 & - & 3.2 & 0.3 & EA & 2024    \\
        MPAE-C~\cite{zou_multiple_2024} & 97.51 & - & 3.7 & 0.3 & EA & 2024    \\
        MPE-NAS~\cite{song_multi-population_2024} & 96.53 & - & 6.4 & 0.78 & EA & 2024    \\
        PEPNAS~\cite{xue_neural_2024} & 97.62 & - & 4.23 & 0.7 & EA & 2024    \\
        SPNAS~\cite{jiang_score_2025} & 98.20 & - & 6.33 & 1.4 & EA & 2025    \\ 
        EmCENAS~\cite{zhang_embedding_2025} & 97.42$\pm$0.03 & - & 4.1 & 0.3 & EA & 2025    \\
        DSGENAS~\cite{xue_graph_2025} & 97.47 & - & 4.8 & 0.5 & EA & 2025    \\
        \hline
        
        MOEA-BUS-S & 98.12$\pm$0.03 & 281 & 5.18 & 1.2 & EA & - \\ 
        MOEA-BUS-M & 98.15$\pm$0.02 & 327 & 6.12 & 1.2 & EA & - \\ 
        MOEA-BUS-L & 98.25$\pm$0.03 & 461 & 7.37 & 1.2 & EA & - \\ 
        MOEA-BUS-XL & \textbf{98.39$\pm$0.03} & 601 & 6.47 & 1.2 & EA & - \\
    \hline\hline
    \end{tabular}
    }
\end{table}
\begin{table*}[h]
    \centering
    \caption{Comparison with state-of-the-art image classifiers on the ImageNet dataset. The search cost excludes the supernet training cost.}
    \label{tab:imagenet_result}
    \scalebox{0.95}{
    \begin{tabular}{c|c|c|c|c|c|c|c}
    \hline\hline
        \textbf{Architecture} & \textbf{Top-1 Acc (\%)} & \textbf{Top-5 Acc (\%)} & \textbf{MAdds (M)} & \textbf{Params (M)} & \textbf{Search Cost (GPU Days)} & \textbf{Search Method} & \textbf{Year} \\
        \hline

        MobileNetV2~\cite{sandler_mobilenetv2_2018} & 72.0 & 91.0 & 300 & 3.4 & - & manual & 2018 \\
        EfficientNet-B0~\cite{tan_efficientnet_2019} & 76.3 & 93.2 & 390 & 5.3 & - & manual & 2019 \\
        \hline

        NASNet-A~\cite{zoph_learning_2018} & 74.0 & 91.6 & 564 & - & 1800 & RL & 2018 \\ 
        MnasNet~\cite{tan_mnasnet_2019} & 76.13 & 92.85 & 391 & 5.2 & - & RL & 2019 \\
        BNAS~\cite{ding_bnas_2021} & 74.3 & 91.5 & - & 3.9 & - & RL & 2021    \\
        DBNAS-B~\cite{yang_deeply_2025} & 75.0 & 92.3 & 385 & 4.4 & 0.9 & RL & 2025    \\
        DBNAS-w/o SE~\cite{yang_deeply_2025} & 77.6 & 93.5 & 386 & 4.9 & 0.9 & RL & 2025    \\
        \hline

        PC-DARTS~\cite{xu_pc-darts_2019} & 75.8 & 92.7 & 597 & 5.3 & 3.8 & GD & 2019 \\
        P-DARTS~\cite{chen_progressive_2019} & 75.6 & 92.6 & 557 & 4.9 & 0.3 & GD & 2019 \\   
        $\beta$-DARTS~\cite{ye_-darts_2022} & 76.1 & 93.0 & 609 & 5.5 & 0.4 & GD & 2022 \\ 
        NAP~\cite{ding_nap_2022} & 75.5 & 92.6 & 574 & 4.8 & 4 & GD & 2022 \\
        EoiNAS~\cite{zhou_exploiting_2022} & 74.4 & 91.7 & 570 & 5.0 & - & GD & 2022    \\
        iDARTS~\cite{wang_idarts_2023} & 75.3 & 92.3 & 568 & 5.1 & 1.9 & GD & 2023    \\
        GENAS~\cite{xue_gradient-guided_2024} & 76.1 & 92.8 & - & 5 & 0.26 & GD & 2024    \\
        SWD-NAS~\cite{xue_self-adaptive_2024} & 75.5 & 92.4 & - & 6.3 & 0.13 & GD & 2024    \\
        DBNAS-A~\cite{yang_deeply_2025} & 74.9 & 92.3 & 382 & 3.7 & 0.6 & GD & 2025    \\
        DBNAS-C~\cite{yang_deeply_2025} & 75.6 & 92.5 & 428 & 4.1 & 0.6 & GD & 2025    \\
        FX-DARTS~\cite{rao_fx-darts_2025} & 76.4 & 93.4 & 610 & 5.1 & 0.17 & GD & 2025    \\
        \hline

        NSGANetV2~\cite{lu_nsganetv2_2020} & 77.4 & 93.5 & 225 & 6.1 & 1 & EA & 2020 \\ 
        CARS~\cite{yang_cars_2020} & 75.2 & 92.5 & 591 & 5.1 & 0.4 & EA & 2020 \\ 
        FairNAS~\cite{chu_fairnas_2021} & 77.5 & - & 392 & - & 12 & EA & 2021 \\
        AutoFormer-Tiny~\cite{chen_autoformer_2021} & 74.7 & 92.6 & 1300 & 5.7 & - & EA & 2021    \\
        AutoFormer-Small~\cite{chen_autoformer_2021} & 81.7 & 95.7 & 5100 & 22.9 & - & EA & 2021    \\
        AutoFormer-Base~\cite{chen_autoformer_2021} & 82.4 & 95.7 & 11000 & 54 & - & EA & 2021    \\
        MixPath~\cite{chu_mixpath_2023} & 77.2 & 93.5 & 378 & 5.1 & 10.3 & EA & 2023 \\
        RelativeNAS~\cite{tan_relativenas_2023} & 75.1 & 92.3 & 563 & 5.1 & - & EA & 2023    \\
        MPAE-A~\cite{zou_multiple_2024} & 74.1 & 91.9 & - & 4.2 & 0.3 & EA & 2024    \\
        MPAE-B~\cite{zou_multiple_2024} & 75.1 & 92.5 & - & 4.8 & 0.3 & EA & 2024    \\
        MPAE-C~\cite{zou_multiple_2024} & 75.7 & 92.7 & - & 5.2 & 0.3 & EA & 2024    \\
        PEPNAS~\cite{xue_neural_2024} & 73.75 & 91.78 & - & 6.71 & 0.7 & EA & 2024    \\
        T-Razor-Tiny~\cite{zhou_training-free_2024} & 75.5 & 92.9 & 1400 & 5.9 & 0.4 & EA & 2024    \\
        T-Razor-Small~\cite{zhou_training-free_2024} & 82.2 & 95.9 & 5100 & 22.3 & 0.4 & EA & 2024    \\
        T-Razor-Base~\cite{zhou_training-free_2024} & 82.3 & 95.6 & 11600 & 53.8 & 0.4 & EA & 2024    \\
        SPNAS~\cite{jiang_score_2025} & 78.62 & 94.07 & 687 & 6.6 & 0.37 & EA & 2025    \\
        HENAS~\cite{jiang_homogeneous_2025} & 78.69 & 94.01 & 580 & - & 0.22 & EA & 2025    \\
        BossNet-S++~\cite{li_bossnas_2025} & 81.4 & 95.6 & 3400 & - & - & EA & 2025    \\
        BossNet-M++~\cite{li_bossnas_2025} & 82.0 & 95.7 & 5800 & - & - & EA & 2025    \\
        BossNet-L++~\cite{li_bossnas_2025} & 83.2 & 96.4 & 10500 & - & - & EA & 2025    \\
        \hline
        
        MOEA-BUS-S & 77.67 & 93.71 & 289 & 6.17 & 0.3 & EA & - \\
        MOEA-BUS-M & 78.28 & 94.04 & 446 & 6.51 & 0.3 & EA & - \\
        MOEA-BUS-L & 78.71 & 94.23 & 461 & 6.62 & 0.3 & EA & - \\
        MOEA-BUS-XL & \textbf{80.03} & 94.42 & 610 & 7.46 & 0.3 & EA & - \\ \hline
        \hline
    \end{tabular}
    }
\end{table*}

\subsection{Results on Standard Datasets}\label{sec:experiment:results}
This section presents and analyzes the experimental results on the CIFAR and ImageNet datasets from different perspectives. We evaluate the performance of the architectures discovered by the proposed algorithm based on accuracy, MAdds, and search time.
Under the available computational resources, to ensure the robustness of our experiments, we conduct multiple rounds of training evaluation on the architectures searched on CIFAR-10 and CIFAR-100, and present the mean and standard deviation of multiple results.
Due to page limitations, the experimental results and analysis on CIFAR-100 are in the \textbf{supplementary material}.
To further validate the performance of MOEA-BUS, we compare the non-dominated architectures obtained by the proposed algorithm with those discovered by other state-of-the-art NAS methods. The selected peer methods can be broadly divided into three categories: manually designed by human experts, EA-based, and non-EA-based (RL-based and GD-based) methods. The results in Tables \ref{tab:cifar10_result} and \ref{tab:imagenet_result} are from the original papers and the classification accuracy, MAdds, number of parameters, and search cost are displayed in them. Among them, the search cost is expressed in GPU days, and the supernet training time is excluded for all of the methods. All results of the proposed method in these tables, i.e., MOEA-BUS-S/M/L/XL, are searched on CIFAR-10 and ImageNet datasets separately and the final four architectures on each dataset are trained.

\textbf{Results on CIFAR-10:} We select four architectures according to different sizes of MAdds, named MOEA-BUS-S/M/L/XL. \tableref{tab:cifar10_result} summarizes the comparison with other state-of-the-art methods. The proposed algorithm achieves the highest average classification accuracy of 98.39\%$\pm$0.03 on the CIFAR-10 dataset. The discovered architecture, MOEA-BUS-S, has a lowest MAdds of 281M among the ENAS methods, and its accuracy of 98.12\% exceeds that of most other methods. FairNAS-A~\cite{chu_fairnas_2021} achieves the accuracy of 98.2\%, comparable to our models, but with significantly higher MAdds of 391M and the search cost of 12 GPU days, highlighting the efficiency of the proposed method.
In recent years, graph neural network (GNN)-guided NAS has emerged as a novel approach~\cite{zhang_embedding_2025,xue_graph_2025}. Compared to these methods, MOEA-BUS maintains significant advantages. 
Compared to gradient-based NAS methods, the proposed approach demonstrates significant advantages in accuracy while maintaining similar MAdds to most of them. Although FX-DARTS~\cite{rao_fx-darts_2025} achieves the lowest MAdds of 195M, its accuracy is significantly inferior to the proposed method. 
In general, the architectures discovered by MOEA-BUS show lower MAdds compared to other methods with the same accuracy, indicating higher computational efficiency.
Besides, compared with other ENAS methods based on multi-population mechanisms, such as MPE-ENAS~\cite{song_multi-population_2024}, the proposed method achieves an advantage of approximately 2\% in classification accuracy.
The total search cost is approximately 1.2 GPU days less than most methods. The performance of different metrics performance surpasses several state-of-the-art NAS methods, demonstrating that the proposed algorithm achieves the best balance between accuracy and computational efficiency.

\textbf{Results on ImageNet:} On the ImageNet dataset, we also provide a set of architectures with different sizes. From \tableref{tab:imagenet_result}, it can be seen that the proposed algorithm demonstrates significant improvements in both classification accuracy and computational efficiency over existing NAS methods. The best architecture discovered by MOEA-BUS achieves a top-1 accuracy of 80.03\% and a top-5 accuracy of 94.42\% with MAdds of 610M. Compared to manually designed architectures like EfficientNet-B0 and B1~\cite{tan_efficientnet_2019}, our architectures provide higher accuracy with competitive or lower computational costs. When compared to other EA-based methods, some architectures have higher accuracy than ours, but the architectures discovered by the proposed algorithm is smaller in MAdds with the least search cost of 0.3 GPU days. 
MixPath~\cite{chu_mixpath_2023}, for instance, achieves a top-1 accuracy of 77.2\% with 378M MAdds and a search cost of 10.3 GPU days, while MOEA-BUS-L surpasses this with a top-1 accuracy of 78.71\% and a search cost of only 0.3 GPU days.
Among all ENAS methods, NAS approaches based on Transformer search space achieve the highest classification accuracy, but result in a significant increase in MAdds. For instance, although BossNet-L++~\cite{li_bossnas_2025} obtains the highest classification accuracy of 83.2\%, its MAdds of 10500M is $17.2\times$ that of MOEA-BUS-XL. T-Razor-Base~\cite{zhou_training-free_2024} achieves a classification accuracy of 82.3\%, but its MAdds of 11000M and parameter count of 54M far exceed the consumption of MOEA-BUS. Compared to AutoFormer-Tiny~\cite{chen_autoformer_2021} and T-Razor-Tiny~\cite{zhou_training-free_2024} with similar parameter scales, their MAdds are still $2\times$ that of MOEA-BUS-XL, and their classification accuracy is significantly lower than MOEA-BUS-XL.
Our architectures also outperform various non-EA-based methods in terms of both accuracy and search efficiency. GENAS~\cite{xue_gradient-guided_2024} achieves a top-1 accuracy of 76.1\% with search cost of 0.26 GPU days, whereas our architectures achieve higher accuracy with comparable search costs.

In summary, the proposed algorithm consistently delivers high-accuracy architectures with lower computational complexity and reduces search cost, outperforming several state-of-the-art NAS methods on three datasets.

\subsection{Ablation Study of Two Key Mechanisms}\label{sec:experiment:abla-key}

The bi-population mechanism in MOEA-BUS algorithm is designed to enhance the exploration and exploitation capabilities during the search process. By allocating different roles to the two populations, the proposed algorithm can explore the search space more comprehensively while simultaneously focusing on high-potential architectures. In the proposed implementation, population 1 focuses on exploring the search space broadly, while population 2 emphasizes exploiting promising regions identified by population 1. This division allows for a more balanced search process, combining the strengths of both exploration and exploitation.

\begin{figure}[!h]
\centering
\subfloat[Search result of NSGA-II (initialized with random sampling)]{
\includegraphics[width=0.45\linewidth]{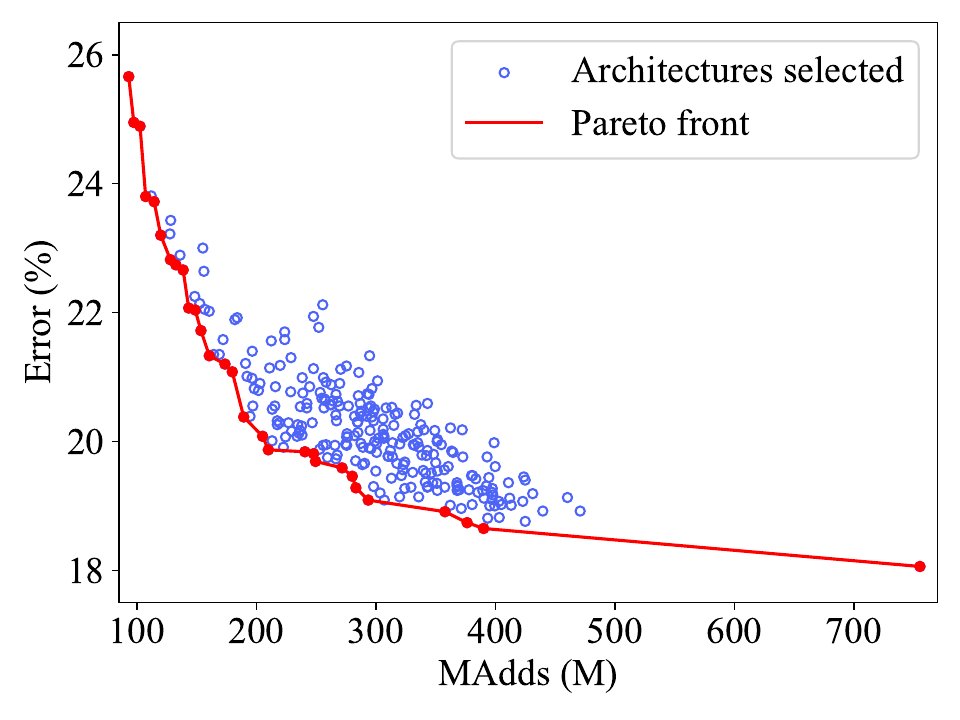}
\label{fig:abla_search_results:wo}
}
\hspace{2mm}
\subfloat[Search result of NSGA-II with multi-population mechanism]{
\includegraphics[width=0.45\linewidth]{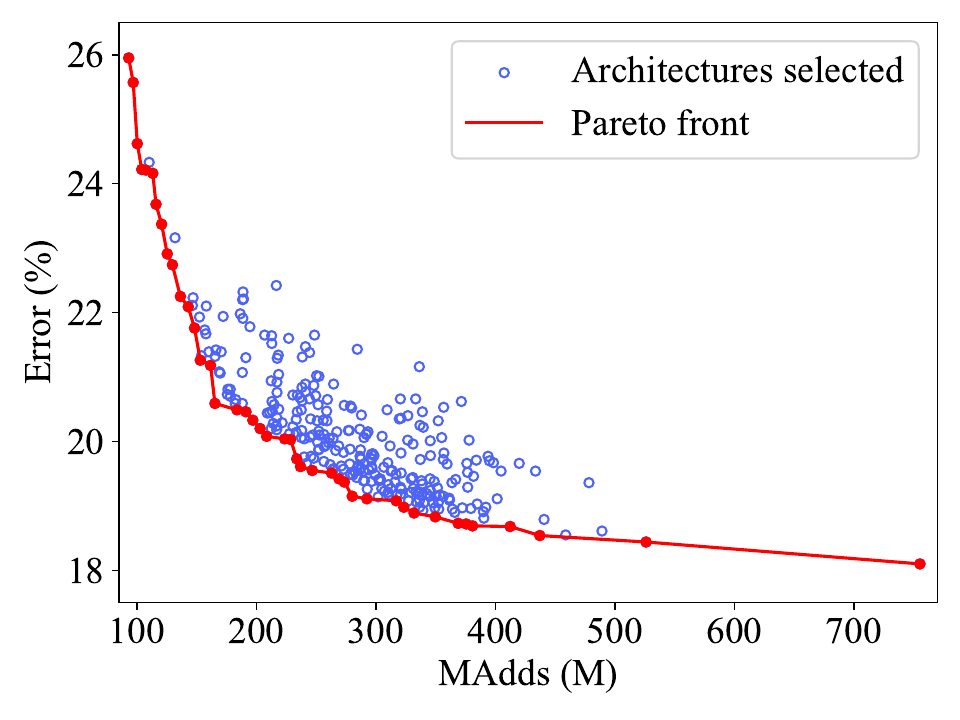}
\label{fig:abla_search_results:bi-population}
}

\subfloat[Search result of NSGA-II (initialized with uniform sampling)]{
\includegraphics[width=0.45\linewidth]{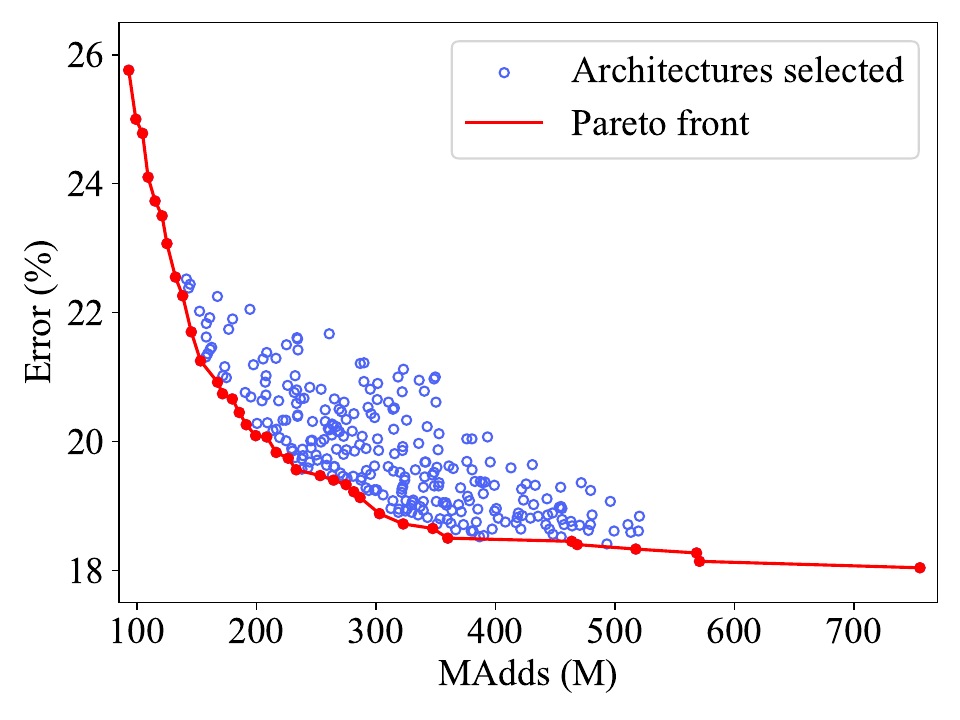}
\label{fig:abla_search_results:uniform-sampling}
}
\hspace{2mm}
\subfloat[Search result of MOEA-BUS]{
\includegraphics[width=0.45\linewidth]{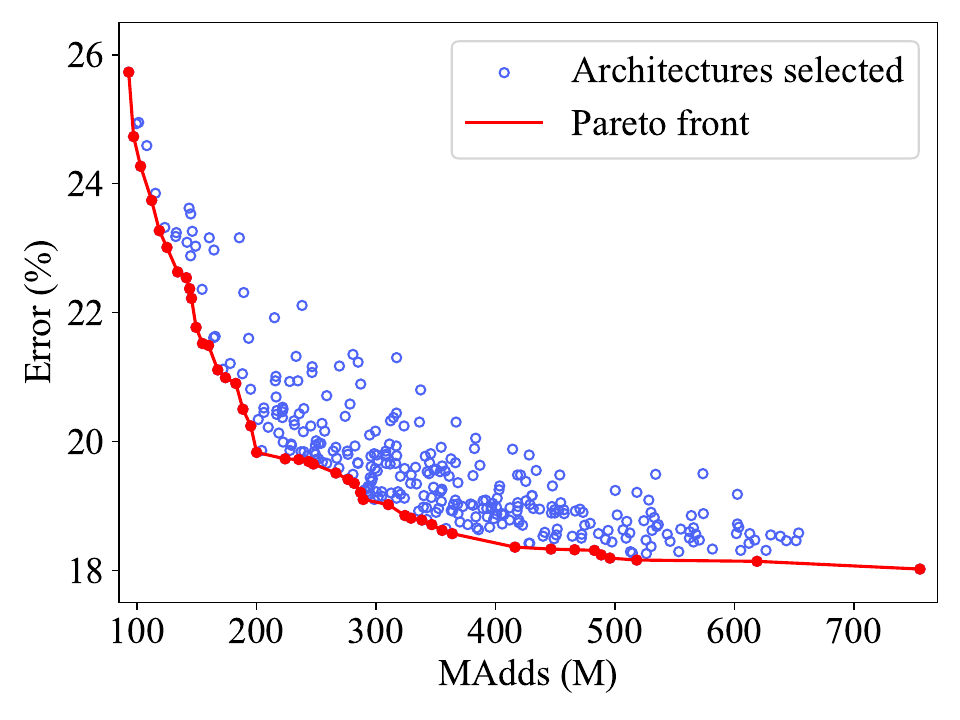}
\label{fig:abla_search_results:MOEA-BUS}
}
\caption{Comparison results of population distribution with and without the proposed two mechanisms on ImageNet.}
\label{fig:abla_search_results}
\end{figure}

\begin{figure}[t]
\centering
\subfloat[Result of architecture distribution entropy (Entropy)]{
\includegraphics[width=0.45\linewidth]{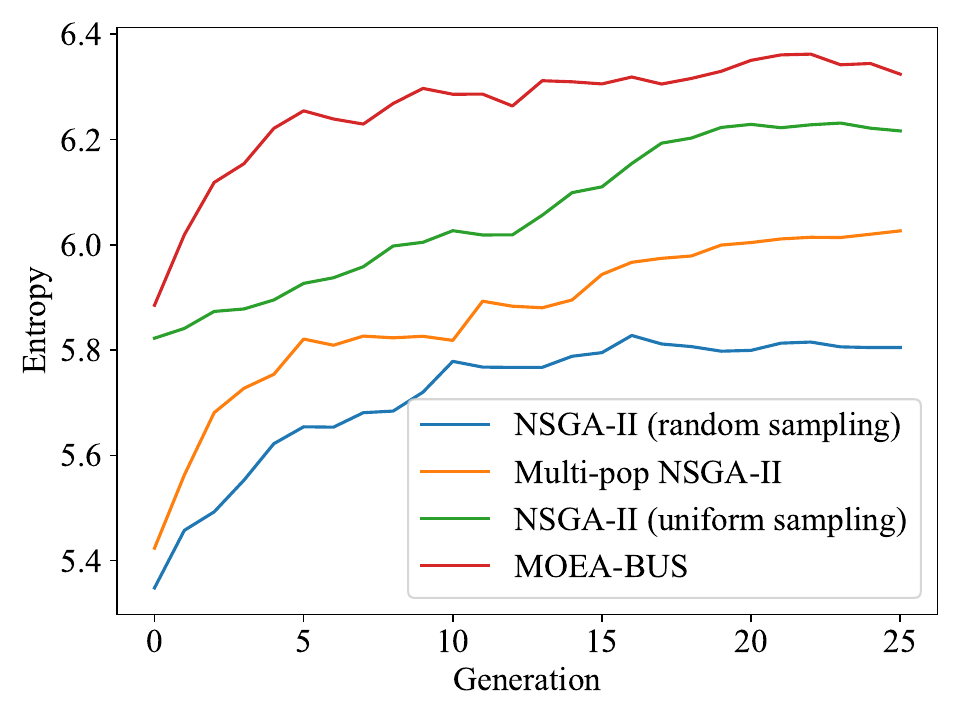}
\label{fig:abla_search_results_fig:Entropy}
}
\hspace{2mm}
\subfloat[Result of hyper-volume (HV)]{
\includegraphics[width=0.45\linewidth]{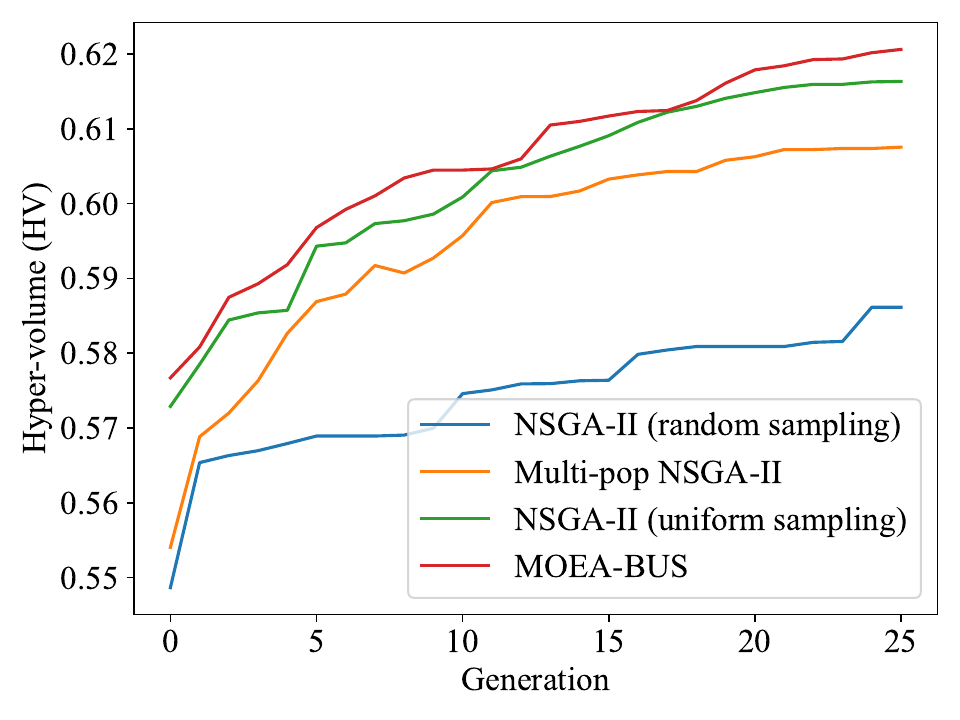}
\label{fig:abla_search_results_fig:hv}
}
\caption{The architecture distribution entropy (Entropy) and hyper-volume (HV) between MOEA-BUS with and without the proposed two mechanisms on ImageNet.}
\label{fig:abla_search_results_fig}
\end{figure}

To further assess the effectiveness of the proposed algorithm, we compare the performance of architectures discovered using the proposed bi-population-based multi-objective algorithm with those obtained using the NSGA-II, a well-known multi-objective evolutionary algorithm.
In the ablation study, we analyze two key mechanisms employed in the paper: the bi-population and uniform sampling methods. Fig. \ref{fig:abla_search_results} contains a total of 4 experimental results, and for each experimental result, we present the final result obtained in the objective space. All points in the figure represent the values of the two search objectives finally retain in archive $\mathcal{A}$, and the red line indicates the Pareto front of the final results.
Furthermore, we design a metric for evaluating the diversity of architecture distributions based on the architecture entropy proposed by Chu et al.~\cite{chu_architecture_2023}, termed architecture distribution entropy: 
\begin{align}
    Entropy \;=\; - \sum_{i,j} p_{ij} \; \log_{2} \bigl(p_{ij}\bigr),\\
    \text{where}\ \ p_{ij} \;=\; \frac{\mathrm{Hist}(a_i', m_j')}{\sum_{i,j} \mathrm{Hist}(a_i', m_j')},
\end{align}
where $a_i'$ and $m_j'$ are normalized accuracy and MAdds, and $\mathrm{Hist}(a_i', m_j')$ is the 2D histogram of normalized accuracy and MAdds. We plot the results from Fig. \ref{fig:abla_search_results} according to search generations, including the proposed architecture distribution entropy and the commonly used hyper-volume (HV) for evaluating population convergence and distribution diversity. The relevant results are presented in Fig. \ref{fig:abla_search_results_fig}.
From Fig. \ref{fig:abla_search_results:wo}, the architectures searched by NSGA-II are heavily concentrated in the MAdds interval of 200M to 400M. This concentration suggests that the results of NSGA-II are limited, focusing predominantly on a narrow region of the search space. As a result, NSGA-II may miss potentially superior architectures in other regions.
The blue line in Fig. \ref{fig:abla_search_results_fig:Entropy} illustrates this point, where it can be observed that this benchmark does not have an advantage in diversity during the initial stage, and Fig. \ref{fig:abla_search_results_fig:hv} reveals that this benchmark has difficulty converging.
When the multi-population mechanism is employed, more architectures with MAdds exceeding 400M are discovered in Fig. \ref{fig:abla_search_results:bi-population}, and as shown in Fig. \ref{fig:abla_search_results_fig:hv}, the two populations can improve search efficiency, enabling more promising architectures to emerge earlier.
Uniform sampling is also applied independently to NSGA-II, achieving richer population diversity. As can be observed from Fig. \ref{fig:abla_search_results:uniform-sampling}, the number of architectures with MAdds less than 200M or greater than 400M increases significantly.
The architectures searched by the proposed algorithm are almost uniformly distributed in the objective space, covering a wide range of MAdds and accuracy values.
From Fig. \ref{fig:abla_search_results_fig:Entropy} and Fig. \ref{fig:abla_search_results_fig:hv}, it can be observed that this ablation setting exhibits good distributional diversity in the early stages of the search and successfully achieves higher diversity and convergence of the final population.
According to Fig. \ref{fig:abla_search_results_fig:Entropy}, the proposed method demonstrates the ability to achieve favorable diversity in the early stages of search and improve throughout the search process rapidly, thereby exploring the entire objective space. Based on Fig. \ref{fig:abla_search_results_fig:hv}, MOEA-BUS can steadily enhance the HV of the entire population, indicating that this method possesses good convergence properties. The population distribution plot of the final results in Fig. \ref{fig:abla_search_results:MOEA-BUS} also substantiates this point, where it can be observed that the points are more uniformly distributed in the objective space compared to other benchmarks, and the Pareto front is also closer to the dominant region.

\subsection{Ablation Study and Analysis of Surrogate Model}\label{sec:experiment:surrogate}
The surrogate model is utilized in this paper to rapidly filter 60,000 architectures, thus the prediction accuracy of the surrogate model is crucial for the results of this experiment. To exclude influences from the search process, we design an experiment focused specifically on the surrogate model. The experimental results are presented in Table \ref{tab:exp_surrogate}, which shows the Kendall's tau correlation coefficient (Ktau) between predicted rankings and real rankings for 1000 architectures under different configurations.

\begin{table}
\centering
\caption{Ablation study on surrogate model prediction performance using four different machine learning models, including the SVM adopted in this paper.}
\label{tab:exp_surrogate}
\begin{tabular}{lcccc} 
\hline
                   & RF     & SVM             & MLP    & Adaboost  \\ 
\hline
Regression+Random  & 0.6257 & 0.6586          & 0.5658 & 0.6748    \\
Regression+Uniform & 0.6807 & 0.7492          & 0.6276 & 0.7228    \\
Pairwise+Random    & 0.6630 & 0.7052          & 0.5731 & 0.7060    \\
Pairwise+Uniform   & 0.6991 & \textbf{0.7721} & 0.6524 & 0.7371    \\
\hline
\end{tabular}
\end{table}

We collect the execution processes from all our previous experiments on the ImageNet dataset, obtaining historical information for approximately 7,000 architectures in total. We perform sampling among these architectures to simulate the impact of different sampling methods on the surrogate model. We sample 1,300 architectures from these candidates, where 300 architectures are used for training the surrogate model and 1000 architectures are employed to evaluate the performance of the surrogate model. We select four commonly used machine learning models as base models: random forest (RF), support vector machine (SVM), multilayer perceptron (MLP), and AdaBoost. In this paper, the proposed surrogate model employs pairwise prediction methods, therefore in our additional experiments, we compare the performance of pairwise prediction with regression prediction. Additionally, we investigate the impact of two initialization strategies: random sampling and uniform sampling.

According to Table \ref{tab:exp_surrogate}, we can intuitively observe the impact of different surrogate patterns (regression prediction and pairwise comparison relationship prediction), initialization methods, and base machine learning models on surrogate performance. It can be observed that both uniform sampling and pairwise prediction can stably enhance the prediction performance of the surrogate model with each base machine learning model. For example, the SVM method used in this paper achieves a 0.0669 Ktau improvement (from 0.7052 to 0.7721) when employing the proposed uniform sampling initialization method combined with pairwise prediction. Comparing the results across different base machine learning models under the Pairwise+Uniform setting, the employed SVM achieves at least a 0.035 Ktau improvement compared to other models (0.7721 vs. 0.7371 for AdaBoost, 0.6991 for RF, and 0.6524 for MLP). Our results demonstrate that the surrogate model employed in the proposed methodology effectively adapts to uniform sampling, enabling accurate performance prediction of candidate architectures.

\subsection{Ablation Study and Analysis of Uniform Sampling}\label{sec:experiment:resultsample}

\begin{figure}[t]
\centering
\subfloat[Distribution of initial architectures with uniform sampling]{
\includegraphics[clip, trim=3.5mm 5mm 3.5mm 3.5mm,width=0.45\linewidth]{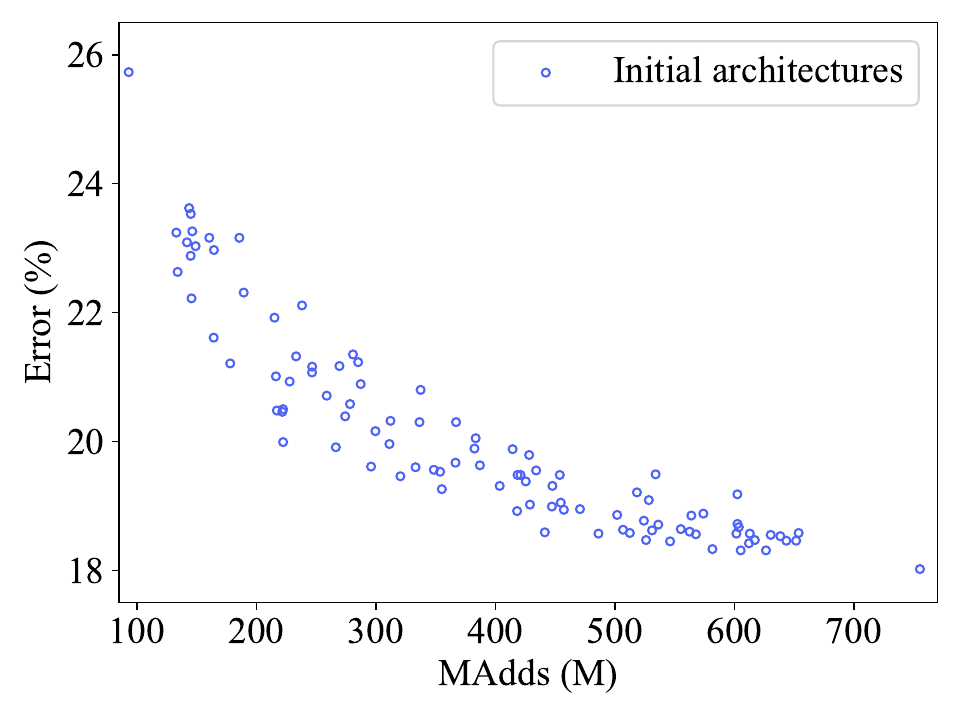}
\label{fig:abla_sampling:uniform}
}
\hspace{2mm}
\subfloat[Distribution of initial architectures with Stratified sampling]{
\includegraphics[clip, trim=3.5mm 5mm 3.5mm 3.5mm,width=0.45\linewidth]{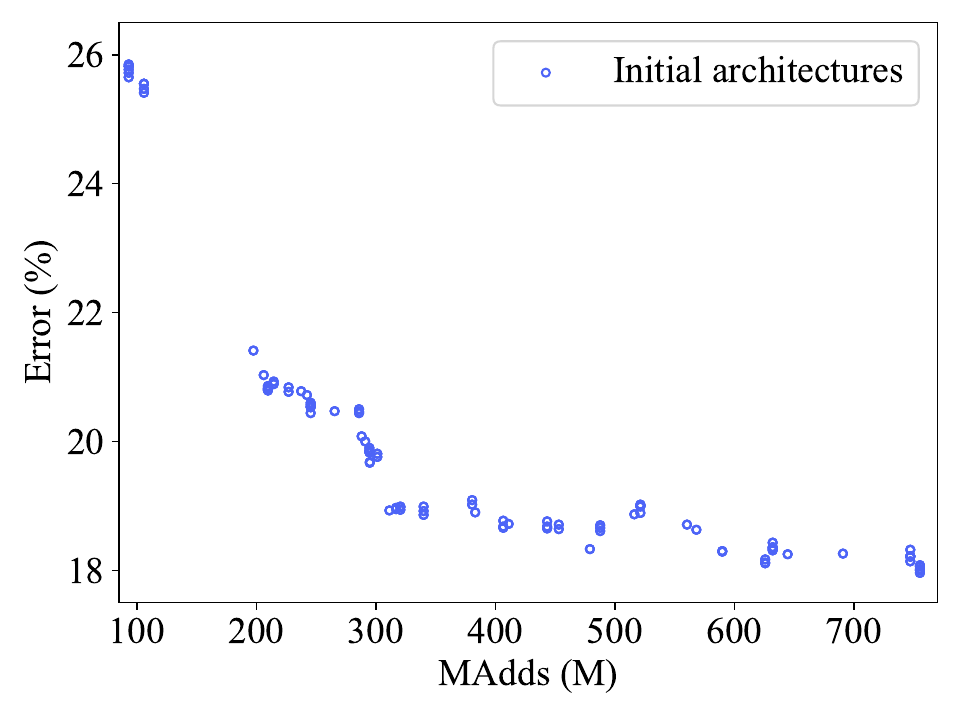}
\label{fig:abla_sampling:stratified}
}

\subfloat[Distribution of initial architectures with Latin Hypercube sampling]{
\includegraphics[clip, trim=3.5mm 5mm 3.5mm 3.5mm,width=0.45\linewidth]{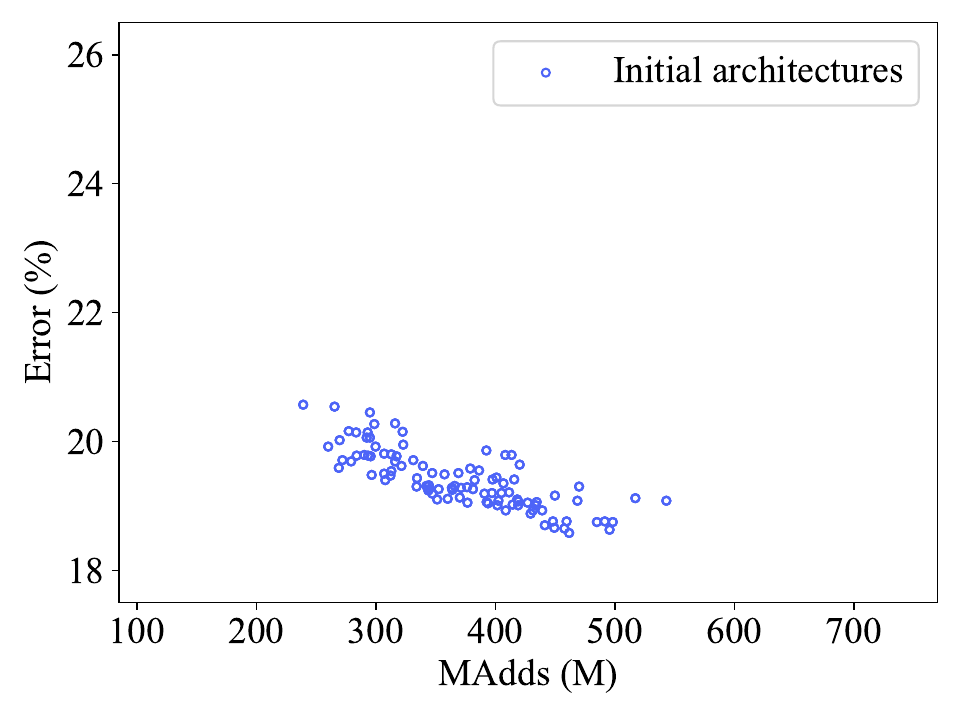}
\label{fig:abla_sampling:hypercube}
}
\hspace{2mm}
\subfloat[Distribution of initial architectures with random sampling]{
\includegraphics[clip, trim=3.5mm 5mm 3.5mm 3.5mm,width=0.45\linewidth]{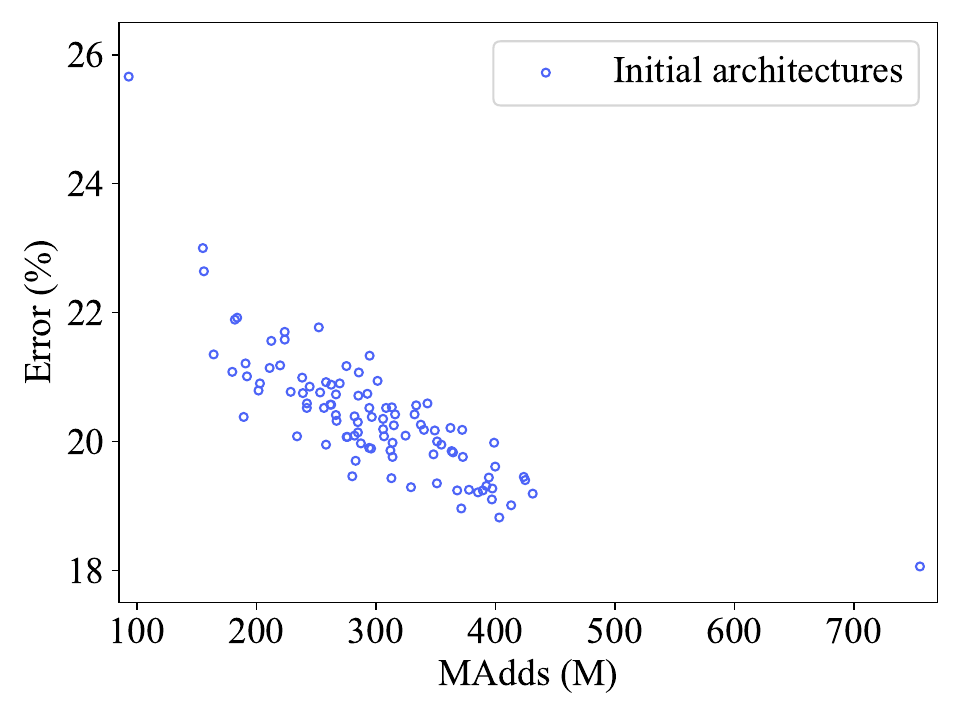}
\label{fig:abla_sampling:random}
}
\caption{Comparison of the distribution of initial architectures using uniform sampling and random sampling.}
\label{fig:abla_sampling}
\end{figure}

\begin{table}[t]
    \centering
    \caption{Architecture distribution entropy and Hyper-volume (HV) of different sampling methods for population initialization.}
    \label{tab:entropy}
\begin{tabular}{l|c|c}
\hline
                            & Entropy & HV     \\
\hline
Uniform sampling (proposed) & \textbf{5.8851}  & \textbf{0.5717} \\
Stratified sampling         & 4.7095  & 0.5553 \\
Latin Hypercube sampling    & 4.6774  & 0.1821 \\
random sampling             & 5.2927  & 0.3314 \\
\hline
\end{tabular}
\end{table}
The uniform sampling employed in the proposed algorithm is designed to ensure that the initial population is distributed as uniformly as possible in the objective space. By uniformly distributing the initial architectures, we avoid the issue of clustering, where architectures are densely populated in certain regions of the search space while other regions remain unexplored. This diversity is crucial for effectively exploring the vast and complex search space of neural architectures, leading to a higher likelihood of discovering optimal solutions. To further assess the effectiveness of the proposed uniform sampling strategy, we compare our results with those obtained using traditional random sampling.
To illustrate this, we plot the effect of using uniform sampling and random sampling in Fig. \ref{fig:abla_sampling}. In the figures, blue dots represent the initial architectures.
We also calculate the architecture distribution entropy and HV for the initial population, with results presented in Table \ref{tab:entropy}.
In Fig. \ref{fig:abla_sampling:uniform}, the initial architectures are almost uniformly distributed in the objective space, which can make the newly generated individuals in later generations to be uniformly spread out, promoting comprehensive exploration of the search space.
The results in Table \ref{tab:entropy} demonstrate that the adopted method can achieve distribution diversity far superior to other methods, and possesses significantly better HV.
In contrast, Fig. \ref{fig:abla_sampling:random} shows that the initial architectures are concentrated in regions with MAdds between 200M and 400M. This concentration limits the search process and reduces the chances of discovering optimal solutions. There are even no architectures in regions with MAdds larger than 450M, which is possible to lead to no larger architectures being generated in the subsequent evolutionary process, neglecting regions with potential for the higher performance. Therefore, architectures derived from traditional random sampling exhibit lower diversity in the initial population, leading to sub-optimal performance during the evolutionary process.
The results in Table \ref{tab:entropy} show that although random sampling can obtain some diversity, it achieves lower HV due to architectures clustering around medium and small sizes, which also makes the subsequent search process challenging.
Additionally, we employ two commonly used sampling methods, stratified sampling method and Latin hypercube sampling method. Due to the close interpolation between the upper and lower bounds of each bit in our encoding scheme, we utilize the encoding approach proposed by Lu\etal for these two sampling methods~\cite{lu_neural_2024}. Fig. \ref{fig:abla_sampling:stratified} presents the results using the stratified sampling method, revealing that this approach, which samples based on the probability distribution of the encoding region, is not suitable for the specific problem addressed in this paper. Although the initial population obtained by the stratified sampling method achieves a favorable distribution in terms of MAdds, it does not exhibit good distribution characteristics from the perspective of the objective space. The entropy and HV values in Table \ref{tab:entropy} also corroborate this observation. Fig. \ref{fig:abla_sampling:hypercube} demonstrates the results using the Latin hypercube sampling method, showing that due to the integer encoding nature of this problem and the still relatively close interpolation between upper and lower bounds, the obtained initial architectures exhibit poor distribution in the objective space.
The substantially lower HV compared to other methods in Table \ref{tab:entropy} illustrates this phenomenon.
Overall, the proposed uniform sampling method ensures a more balanced exploration, resulting in higher-quality architectures with better trade-offs between accuracy and computational complexity.

\subsection{Ablation Study and Analysis of Bi-population Mechanism}\label{sec:experiment:Bi-population}
\begin{figure}[t]
\centering
\subfloat[Population 1 and population 2 share elite individuals with each other]{
\includegraphics[width=0.45\linewidth]{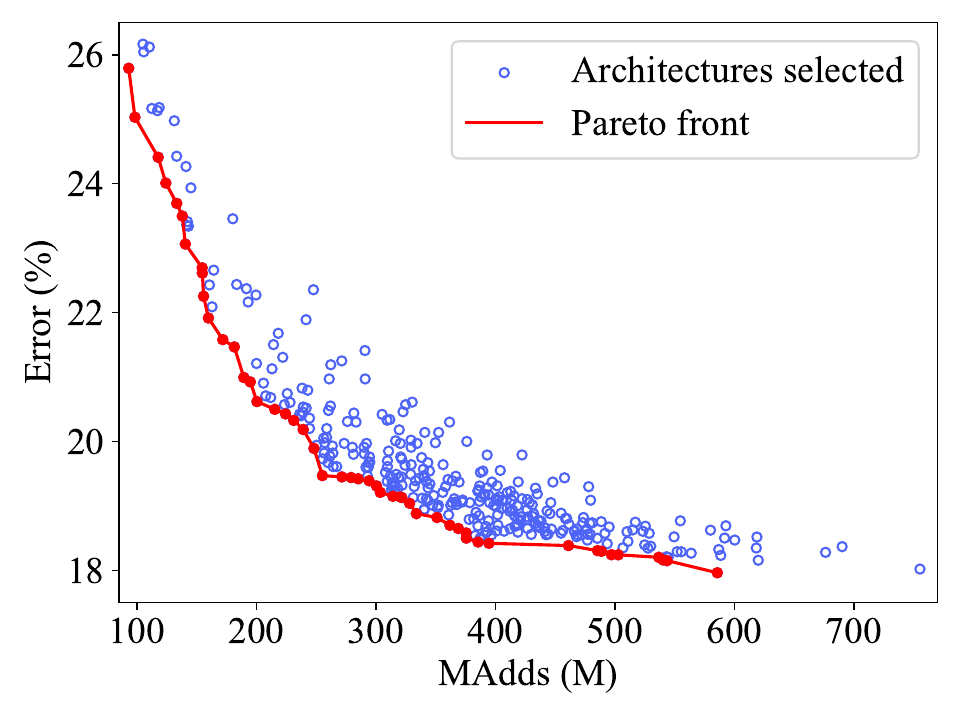}
\label{fig:abla_exchange:both}
}
\hspace{2mm}
\subfloat[Population 1 shares elite individuals with population 2]{
\includegraphics[width=0.45\linewidth]{figs/abla_pareto/proposed.pdf}
\label{fig:abla_exchange:proposed}
}
\caption{Comparison results of population distribution with different exchange rules.}
\label{fig:abla_exchange}
\end{figure}

In this paper, the bi-population mechanism is designed to enhance population diversity and improve search efficiency. To investigate the impact of individual exchange rules in the bi-population mechanism, we design a comparative experiment. The results are shown in Figure \ref{fig:abla_exchange}. In this comparative experiment, elite individuals from two populations are exchanged with each other. This means that, unlike MOEA-BUS method, elite individuals from population 2 are allowed to enter population 1. We plot the results of this comparative experiment. As can be observed from Figure \ref{fig:abla_exchange:both}, in regions where MAdds is less than 250M and greater than 450M, the comparative experiment obtained fewer individuals than the proposed method. Additionally, the Pareto front of the comparative experiment is also significantly worse than the proposed method, achieving architectures with higher error rates under the same MAdds. We believe that medium-sized architectures in population 1 are more inclined to achieve non-dominated frontier positions, thereby preventing population 1 from effectively focusing on extreme architectures.
Additionally, we calculate the architecture distribution entropy and HV for both approaches. The final population obtained by MOEA-BUS achieves architecture entropy of 6.32 and an HV of 0.62. When the two populations exchange with each other, the final population exhibits architecture entropy of 6.07 and an HV of 0.60. These metrics also demonstrate that population diversity decreases when the two populations exchange with each other.

To further analyze the bi-population mechanism of MOEA-BUS, we perform a series of experiments by varying the initial population size for the sub-search process and present the results in \tableref{tab:parasize}.
First, we maintain the total number of 100 individuals unchanged and altered the ratio between population 1 and population 2, denoted as $P_1$ and $P_2$. In the proposed method, we employ parameter values of $P_1=25$ and $P_2=75$, denoted as the parameter pair (25, 75). We design two other sets of experiments, including (50, 50) and (75, 25). According to the results in \tableref{tab:parasize}, we found that individuals with medium-sized architectures achieve lower accuracy compared to the proposed configuration. Moreover, the results under the (75, 25) setting are dominated by those of (25, 75) and (50, 50). This indicates that maintaining a larger number of individuals in population 2 is beneficial for medium-sized architectures. This inference aligns with the quantity distribution in the search space. As illustrated in \figref{fig:random5000}, the number of medium-sized architectures is evidently far greater than that of extreme architectures, thus necessitating the maintenance of a larger population 2.
Furthermore, we experiment with different sizes for $P_1$ and $P_2$ with the same ratio but a larger total number of individuals. Three sets of parameter values for $P_1$ and $P_2$ are employed, and the sizes are varied as follows: (15, 45), (35, 105), and (45, 135) marked with ``$^*$''. Increasing the population size generally improves the diversity of the architectures and leads to better performance in terms of accuracy. However, larger populations also increase computational costs. \tableref{tab:parasize} also summarizes the results of our ablation study on different population sizes. A population size of (45, 135) gives slightly better results than (25, 75), but the performance improvement is very limited when considering the higher computational cost. When the population size is (15, 45), although this configuration cost the least computational search time, the architectures with lower accuracy are obtained compared to larger populations. The limited diversity in the smaller population size restricts the exploration in the search space, resulting in sub-optimal architectures. We find that the population size of (25, 75) strikes an optimal balance between performance and computational efficiency.
To obtain search results with the same number of actual evaluations, we modify the number of search generations for these three experimental settings. Specifically, for (15, 45), we increase generations by four; for (35, 105), we decrease generations by four; for (45, 135), we decrease generations by eight. It can be observed that although increasing the initial population size can reduce the overall search time, it yields worse architectures at medium scales. While reducing the initial population can obtain more search generations, the training data for the surrogate model is acquired at a slower rate, resulting in increased search duration without achieving better results. The ablation experiments in Table \ref{tab:parasize} demonstrate the rationality of the employed setting (25, 75).

\begin{table}[h]
    \centering
    \caption{Comparison results on ImageNet with different sizes/ratios of the two initial populations. }
    \label{tab:parasize}
    \begin{tabular}{c|c|c|c|c}
         \hline
         \begin{tabular}{@{}c@{}}\textbf{Size}\\\textbf{($P_1$, $P_2$)}\end{tabular} & \begin{tabular}{@{}c@{}}\textbf{Top-1}\\\textbf{Acc (\%)}\end{tabular}& \begin{tabular}{@{}c@{}}\textbf{MAdds}\\\textbf{(M)}\end{tabular} & \begin{tabular}{@{}c@{}}\textbf{Search Cost}\\\textbf{(GPU Days)}\end{tabular} & \begin{tabular}{@{}c@{}}\textbf{Number of real}\\\textbf{evaluations}\end{tabular} \\ \hline
        (25, 75)  & 78.71 & 461 & 0.30 & 350 \\ \hline
        (50, 50)  & 78.44 & 447 & 0.30 & 350 \\ \hline
        (75, 25)  & 78.38 & 464 & 0.29 & 350 \\ \hline \hline
        (15, 45)  & 78.35 & 501 & 0.40 & 350 \\ \hline
        (15, 45)$^*$  & 78.20 & 489 & 0.25 & 310 \\ \hline
        (35, 105) & 78.42 & 445 & 0.28 & 350 \\ \hline
        (35, 105)$^*$ & 78.73 & 475 & 0.37 & 390 \\ \hline
        (45, 135) & 78.15 & 438 & 0.25 & 350 \\
        \hline
        (45, 135)$^*$ & 78.78 & 503 & 0.44 & 430 \\
        \hline
    \end{tabular}
\end{table}

\section{Conclusion and Future Work}\label{sec:conclusion}
In this paper, we present MOEA-BUS, a multi-objective evolutionary algorithm based on bi-population with uniform sampling for neural architecture search. The proposed method aims to address the challenges of generating high-performance neural architectures while balancing computational complexity. By integrating a uniform sampling strategy for initializing the population and a bi-population mechanism for evolutionary search, MOEA-BUS effectively explores the search space, ensures diversity and optimizing multiple objectives. We validate the effectiveness of MOEA-BUS on three image classification datasets: CIFAR-10, CIFAR-100, and ImageNet. Our experimental results demonstrate that MOEA-BUS outperforms several state-of-the-art NAS methods in terms of accuracy and computational efficiency.

While this study utilized the MobileNetV3 backbone to align with common mobile deployment scenarios and ensure fair comparison with recent benchmarks, we acknowledge that evaluating a single search space is a limitation. However, the core contributions of this work—specifically the uniform sampling and bi-population mechanisms—are designed to be architecture-agnostic.

Future work will aim to demonstrate this broad applicability by extending the framework to other backbone architectures, such as ResNet, EfficientNet, and Vision Transformers. Additionally, we plan to adapt the proposed framework to other domains beyond image classification, including semantic segmentation and object detection, by designing corresponding search operators and surrogate models. These efforts, combined with comprehensive cross‑task benchmarks, will further validate the generality and robustness of the proposed strategies.

\bibliography{cite}
\bibliographystyle{ieeetr}

\vspace{-20 pt}

\begin{IEEEbiography}
[{\includegraphics[width=1in,height=1.25in,clip,keepaspectratio]{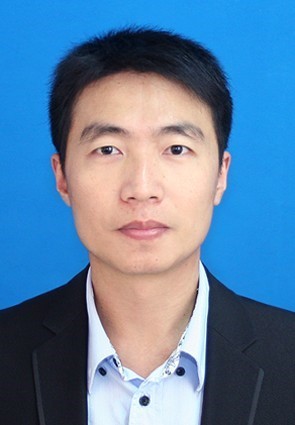}}]
{Yu Xue} (Senior Member, IEEE) received the Ph.D. degree from the School of Computer Science and Technology, Nanjing University of Aeronautics and Astronautics, Nanjing, China, in 2013. He was a Visiting Scholar with the School of Engineering and Computer Science, Victoria University of Wellington, Wellington, New Zealand, from August 2016 to August 2017. He was a Research Scholar with the Department of Computer Science and Engineering, Michigan State University, East Lansing, MI, USA, from October 2017 to November 2018. He is currently a Professor with the School of Software, Nanjing University of Information Science and Technology, Nanjing. His research interests include deep learning, evolutionary computation, machine learning, computer vision, and feature map selection.
\end{IEEEbiography}

\vspace{-20 pt}

\begin{IEEEbiography}
[{\includegraphics[width=1in,height=1.25in,clip,keepaspectratio]{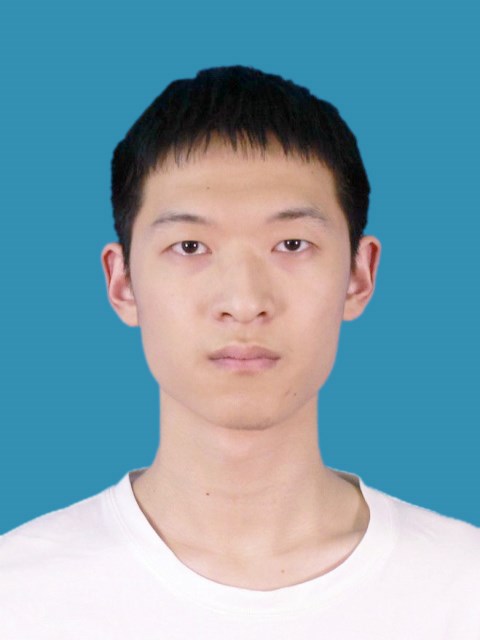}}]
{Pengcheng Jiang} (Graduate Student Member, IEEE) received the B.E. degree from Nanjing University of Information Science and Technology, China, in 2020. He is currently pursuing the Ph.D. degree with the School of Software in Nanjing University of Information Science and Technology, China. His current research interests include feature selection, evolutionary computation, neural architecture search, and model compression.
\end{IEEEbiography}

\vspace{-20 pt}

\begin{IEEEbiography}
[{\includegraphics[width=1in,height=1.25in,clip,keepaspectratio]{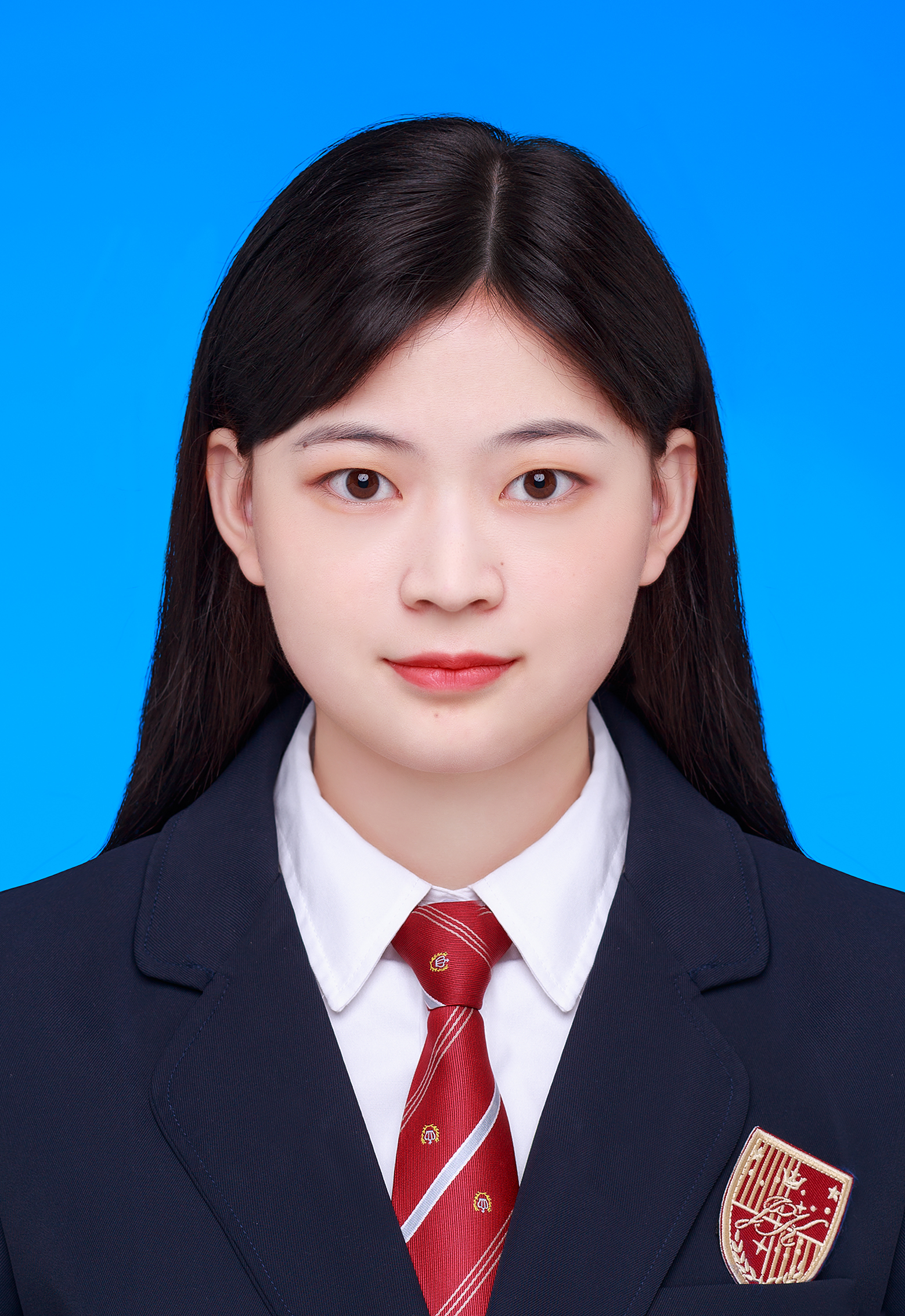}}]
{Chenchen Zhu} received the B.E. degree from Nanjing University of Information Science and Technology, China, in 2022, where she is currently pursuing a master’s degree. Her research interests include deep learning, multi-objective optimization and neural architecture search.
\end{IEEEbiography}

\vspace{-20 pt}

\begin{IEEEbiography}
[{\includegraphics[width=1in,height=1.25in,clip,keepaspectratio]{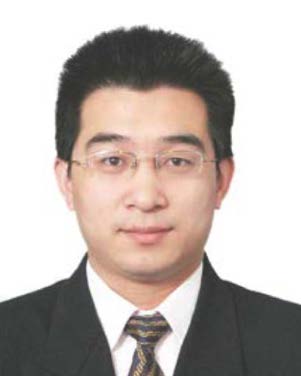}}]
{Yong Zhang} (Senior Member, IEEE) received the Ph.D. degree in control theory and control engineering from China University of Mining and Technology, Xuzhou, China, in 2009. He is currently a Professor at the School of Information and Control Engineering, China University of Mining and Technology. His research interests cover swarm intelligence and machine learning.
\end{IEEEbiography}

\vspace{-20 pt}

\begin{IEEEbiography}[{\includegraphics[width=1in,height=1.25in,clip,keepaspectratio]{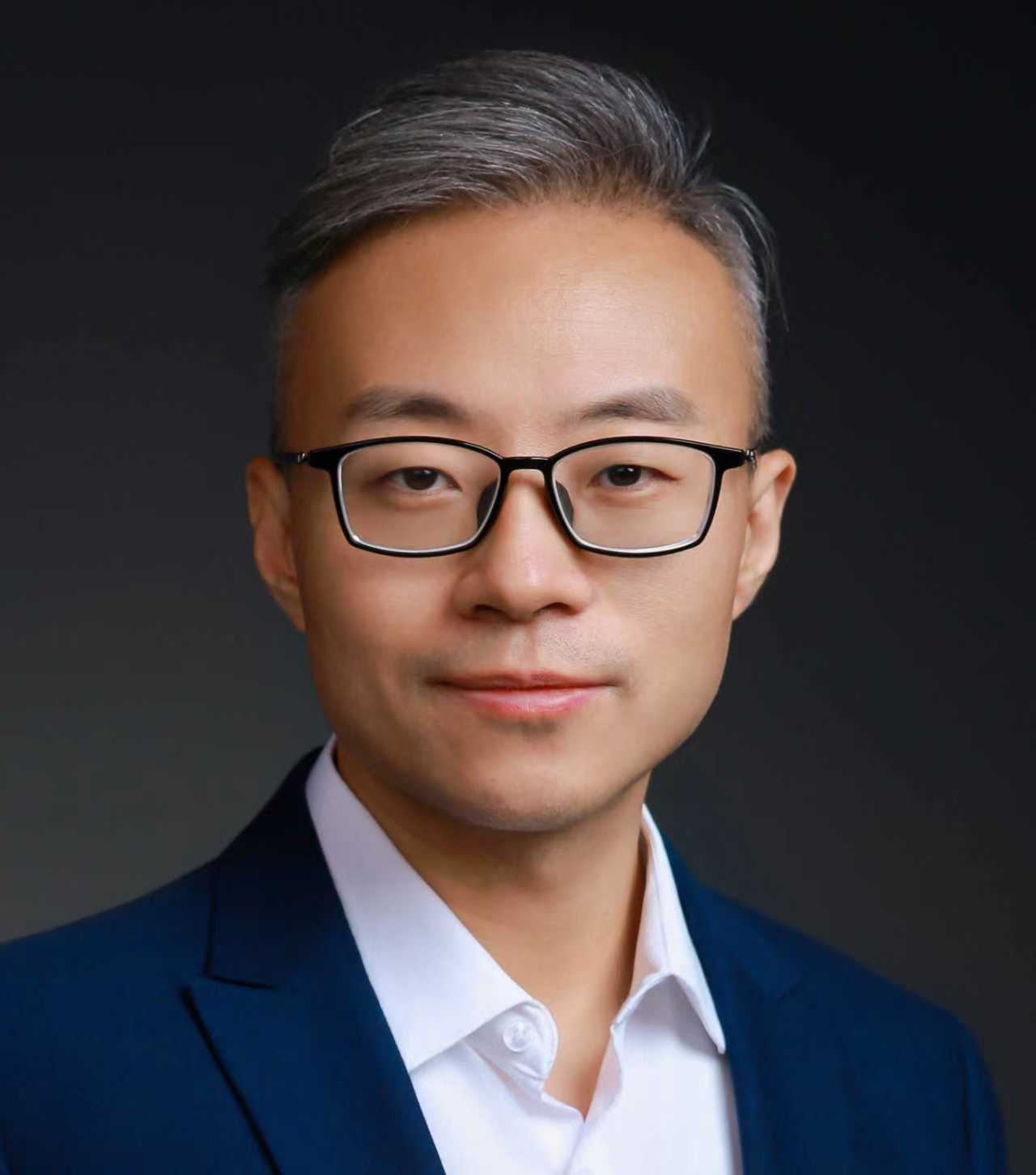}}]
{Ran Cheng} (Senior Member, IEEE)
received the B.Sc. degree from the Northeastern University, Shenyang, China, in 2010, and the Ph.D. degree from the University of Surrey, Guildford, U.K., in 2016. 
He is currently an Associate Professor with the Department of Data Science and Artificial Intelligence, and the Department of Computing, The Hong Kong Polytechnic University, Hong Kong SAR, China. 
He is a recipient of the IEEE Transactions on Evolutionary Computation Outstanding Paper Award (2018 and 2021), the IEEE Computational Intelligence Society Outstanding Ph.D. Dissertation Award (2019), the IEEE Computational Intelligence Magazine Outstanding Paper Award (2020), and the IEEE Computational Intelligence Society Early Career Award (2025). 
He is the Founding Chair of the IEEE Computational Intelligence Society Shenzhen Chapter. 
He is an Associate Editor of IEEE Transactions on Evolutionary Computation, IEEE Transactions on Artificial Intelligence, IEEE Transactions on Emerging Topics in Computational Intelligence, and IEEE Transactions on Cognitive and Developmental Systems.
\end{IEEEbiography}

\vspace{-20 pt}

\begin{IEEEbiography}
[{\includegraphics[width=1in,height=1.25in,clip,keepaspectratio]{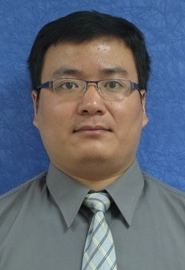}}]
{Kaizhou Gao} (Senior Member, IEEE) received the B.Sc. degree from Liaocheng University, Liaocheng, China, in 2005, the master’s degree from Yangzhou University, Yangzhou, China, in 2008, and the Ph.D. degree from Nanyang Technological University (NTU), Singapore, in 2016. He is currently an Associate Professor with the Macau Institute of Systems Engineering, Macau University of Science and Technology. He has published over 100 refereed papers. His research interests include intelligent computation, optimization, scheduling, and intelligent transportation. He is an Associate Editor of IEEE Transactions on Intelligent Transportation Systems, Swarm and Evolutionary Computation, and Expert Systems with Applications.
\end{IEEEbiography}

\vspace{-20 pt}

\begin{IEEEbiography}
[{\includegraphics[width=1in,height=1.25in,clip,keepaspectratio]{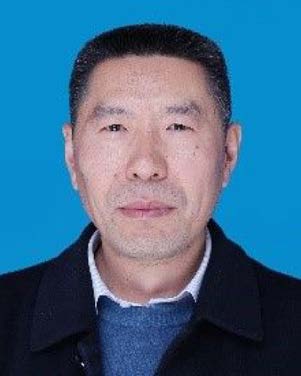}}]
{Dunwei Gong} (Senior Member, IEEE) is a professor and the Dean of the School of Microelectronics at Qingdao University of Science and Technology. He was selected as a Shandong Taishan Scholar Distinguished Expert in 2025, and a Clarivate Highly Cited Researcher in 2022 and 2023. His main research interests include intelligent optimization theory and applications. He investigated seven National Natural Science Foundation of China (including one key project) and one National Key R\&D Program project, received five Prizes for Natural Science, and published 75 papers in CAS Zone I journals, accumulating 10824 Web of Science citations and an H-index of 56.
\end{IEEEbiography}

\includepdf[pages=-, fitpaper=true]{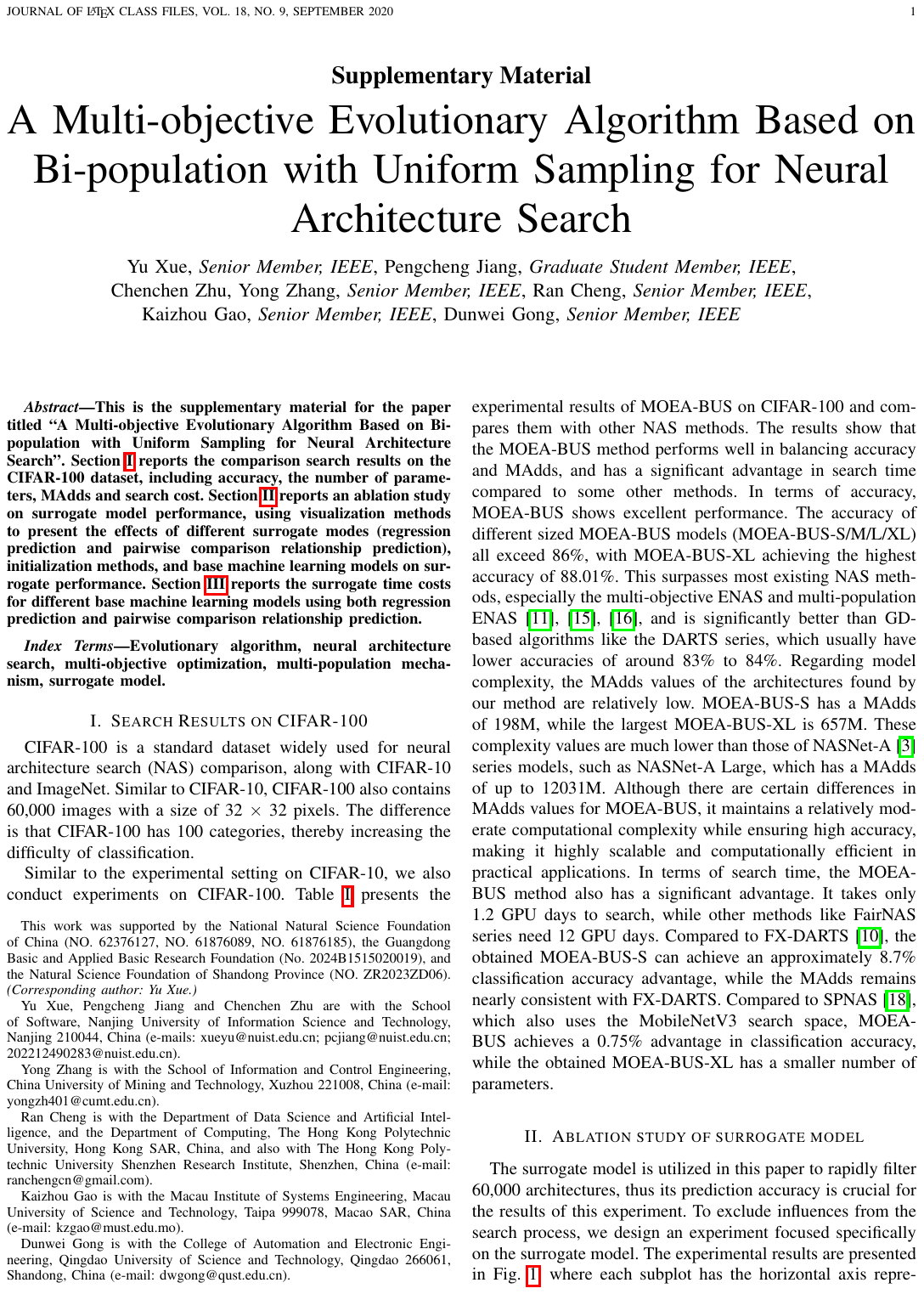}

\end{document}